\documentclass[10pt,twocolumn,letterpaper]{article}

\usepackage[pagenumbers]{cvpr} %

\usepackage{graphicx}
\usepackage{amsmath}
\usepackage{amssymb}
\usepackage{booktabs}

\usepackage[export]{adjustbox}
\usepackage{placeins}
\usepackage{graphbox} %
\usepackage{mathtools}

\usepackage[pagebackref,breaklinks,colorlinks]{hyperref}

\usepackage[capitalize]{cleveref}
\crefname{section}{Sec.}{Secs.}
\Crefname{section}{Section}{Sections}
\Crefname{table}{Table}{Tables}
\crefname{table}{Tab.}{Tabs.}

\def\cvprPaperID{56} %
\def\confName{CVPR}
\def\confYear{2022}

\providecommand{\imwidth}{}
\providecommand{\halfimwidth}{}
\providecommand{\impath}[1]{}
\providecommand{\davisseq}{}

\newcommand{\figmodel}{
  \begin{figure*}[t]
    \centering
    \includegraphics[width=1.0\textwidth]{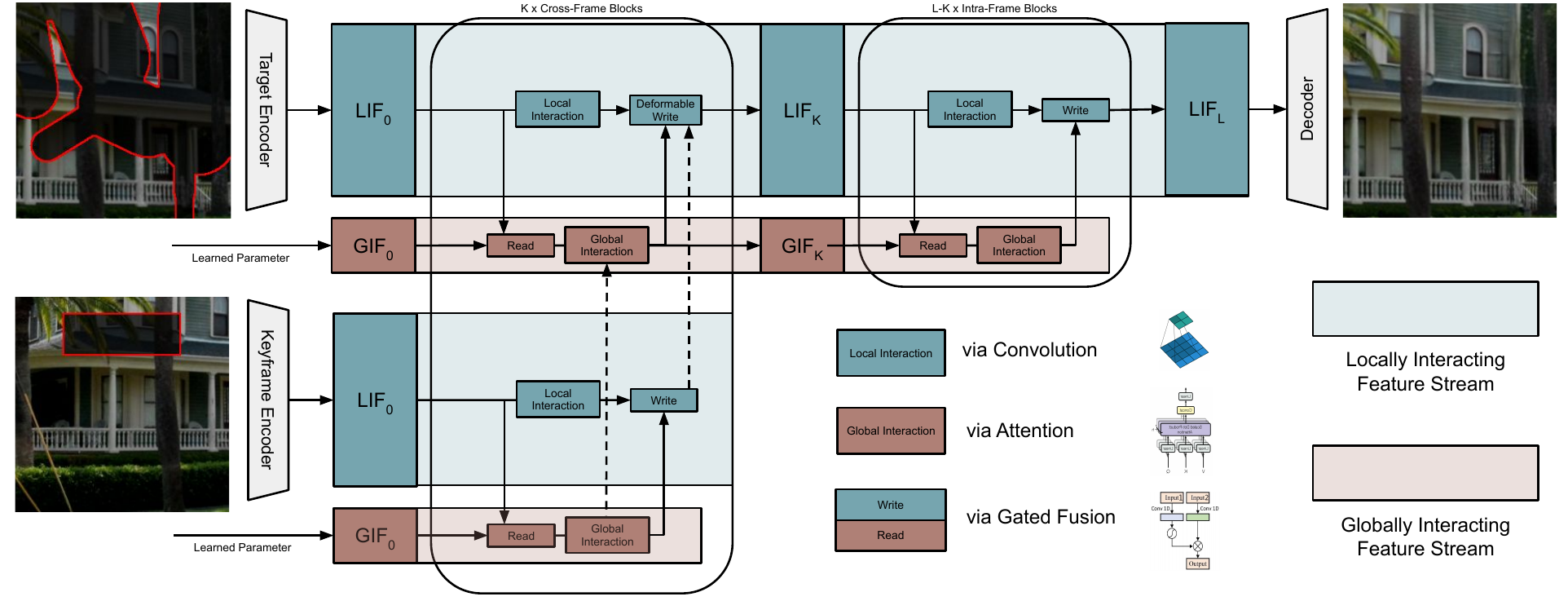}
  \caption{\label{fig:model}
    Overview of our model operating with two streams: A locally interacting
    feature (LIF) stream, relying on convolutions for high-frequency modeling,
    and a globally interacting feature (GIF) stream, relying on attention for
    low-frequency modeling.%
    \vspace{-1.5em}
  }
  \end{figure*}
}
\newcommand{\figone}{
    \begin{center}
        \includegraphics[width=1.00\textwidth, trim=0em 0em 0em 0em, clip]{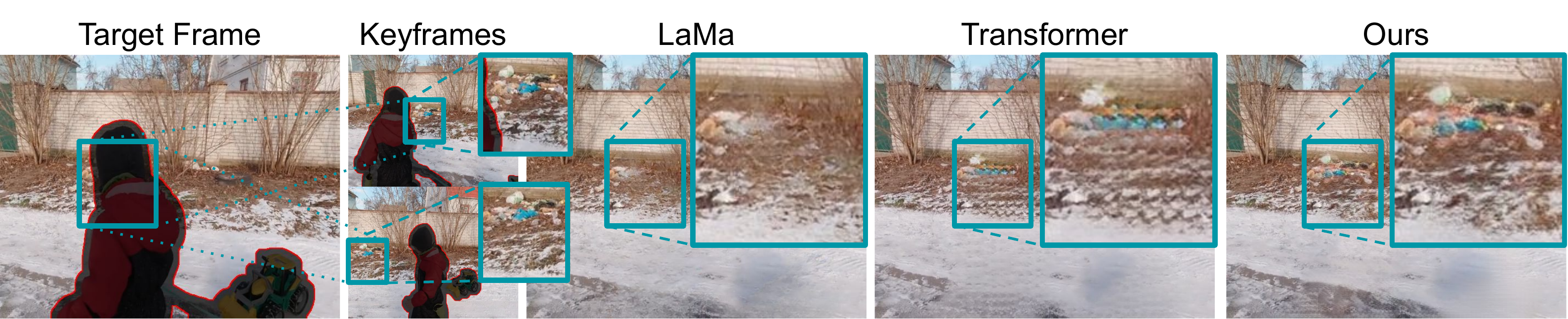}%
        \vspace{-0.5em}%
        \captionof{figure}{Single image inpainting approaches such as LaMa\cite{lama} (third col.) cannot propagate context from keyframes (second col.) to a target frame (first col.). By aggregating features globally across frames, transformer-based approaches (fourth col.) can propagate coarse context about a blue object that is visible in the keyframes but not in the target frame. However, it fails to propagate high-frequency details about object locations and textures, resulting in repetitive patterns and artifacts. By modeling both local and global interactions within and across frames, our approach (last col.) successfully propagates high-frequency context and accurately reconstructs the background.%
      }
      \label{figone}
    \end{center}
  \vspace{0.5em}
}

\newcommand{\figqualinpaint}{
\begin{figure}
\centering
\scriptsize
	\renewcommand{\imwidth}{0.075\textwidth}
	\renewcommand{\halfimwidth}{0.0375\textwidth}
  \renewcommand{\impath}[1]{img/##1} 
	\setlength{\tabcolsep}{0pt}
	\begin{tabular}{c@{\hskip 2pt}c@{\hskip 2pt}cccc}
	\toprule
    keyframes & target & Aligned & LaMa & Transformer & Ours \\
	\midrule
   \includegraphics[align=c,width=\halfimwidth]{\impath{ours/000000_src}} &
   \includegraphics[align=c,width=\imwidth]{\impath{ours/000000_masked_image}} &
   \includegraphics[align=c,width=\imwidth]{\impath{jinx/000000_explicit}} &
   \includegraphics[align=c,width=\imwidth]{\impath{lama/000000_predicted_image}} &
   \includegraphics[align=c,width=\imwidth]{\impath{transformer/000000_predicted_image}} &
   \includegraphics[align=c,width=\imwidth]{\impath{ours/000000_predicted_image}} \\
   \includegraphics[align=c,width=\halfimwidth]{\impath{ours/000002_src}} &
   \includegraphics[align=c,width=\imwidth]{\impath{ours/000002_masked_image}} &
   \includegraphics[align=c,width=\imwidth]{\impath{jinx/000002_explicit}} &
   \includegraphics[align=c,width=\imwidth]{\impath{lama/000002_predicted_image}} &
   \includegraphics[align=c,width=\imwidth]{\impath{transformer/000002_predicted_image}} &
   \includegraphics[align=c,width=\imwidth]{\impath{ours/000002_predicted_image}} \\
   \includegraphics[align=c,width=\halfimwidth]{\impath{ours/000008_src}} &
       \includegraphics[align=c,width=\imwidth]{\impath{ours/000008_masked_image}} &
       \includegraphics[align=c,width=\imwidth]{\impath{jinx/000008_explicit}} &
       \includegraphics[align=c,width=\imwidth]{\impath{lama/000008_predicted_image}} &
\includegraphics[align=c,width=\imwidth]{\impath{transformer/000008_predicted_image}} &
       \includegraphics[align=c,width=\imwidth]{\impath{ours/000008_predicted_image}} \\
   \includegraphics[align=c,width=\halfimwidth]{\impath{ours/000013_src}} &
       \includegraphics[align=c,width=\imwidth]{\impath{ours/000013_masked_image}} &
       \includegraphics[align=c,width=\imwidth]{\impath{jinx/000013_explicit}} &
       \includegraphics[align=c,width=\imwidth]{\impath{lama/000013_predicted_image}} &
\includegraphics[align=c,width=\imwidth]{\impath{transformer/000013_predicted_image}} &
       \includegraphics[align=c,width=\imwidth]{\impath{ours/000013_predicted_image}} \\
   \includegraphics[align=c,width=\halfimwidth]{\impath{ours/000036_src}} &
       \includegraphics[align=c,width=\imwidth]{\impath{ours/000036_masked_image}} &
       \includegraphics[align=c,width=\imwidth]{\impath{jinx/000036_explicit}} &
       \includegraphics[align=c,width=\imwidth]{\impath{lama/000036_predicted_image}} &
\includegraphics[align=c,width=\imwidth]{\impath{transformer/000036_predicted_image}} &
       \includegraphics[align=c,width=\imwidth]{\impath{ours/000036_predicted_image}} \\
	\bottomrule
  \end{tabular}\vspace{-1.25em}
\caption{\label{figqualinpaint} Qualitative results on guided image
  inpainting.\vspace{-1.5em}}
\end{figure}
}

\newcommand{\figqualinpaintsupp}{
\begin{figure*}[b]
\centering
\scriptsize
	\renewcommand{\imwidth}{0.19\textwidth}
	\renewcommand{\halfimwidth}{0.095\textwidth}
  \renewcommand{\impath}[1]{img/##1} 
	\setlength{\tabcolsep}{0pt}
	\begin{tabular}{c@{\hskip 2pt}c@{\hskip 2pt}c@{\hskip 2pt}ccc}
	\toprule
    keyframes & target & Aligned & LaMa & Transformer & Ours \\
	\midrule
   \includegraphics[align=c,width=\halfimwidth]{\impath{ours/000001_src}} &
   \includegraphics[align=c,width=\imwidth]{\impath{ours/000001_masked_image}} &
   \includegraphics[align=c,width=\imwidth]{\impath{jinx/000001_explicit}} &
   \includegraphics[align=c,width=\imwidth]{\impath{lama/000001_predicted_image}} &
   \includegraphics[align=c,width=\imwidth]{\impath{transformer/000001_predicted_image}} &
   \includegraphics[align=c,width=\imwidth]{\impath{ours/000001_predicted_image}} \\
   \includegraphics[align=c,width=\halfimwidth]{\impath{ours/000003_src}} &
   \includegraphics[align=c,width=\imwidth]{\impath{ours/000003_masked_image}} &
   \includegraphics[align=c,width=\imwidth]{\impath{jinx/000003_explicit}} &
   \includegraphics[align=c,width=\imwidth]{\impath{lama/000003_predicted_image}} &
   \includegraphics[align=c,width=\imwidth]{\impath{transformer/000003_predicted_image}} &
   \includegraphics[align=c,width=\imwidth]{\impath{ours/000003_predicted_image}} \\
   \includegraphics[align=c,width=\halfimwidth]{\impath{ours/000010_src}} &
       \includegraphics[align=c,width=\imwidth]{\impath{ours/000010_masked_image}} &
       \includegraphics[align=c,width=\imwidth]{\impath{jinx/000010_explicit}} &
       \includegraphics[align=c,width=\imwidth]{\impath{lama/000010_predicted_image}} &
\includegraphics[align=c,width=\imwidth]{\impath{transformer/000010_predicted_image}} &
       \includegraphics[align=c,width=\imwidth]{\impath{ours/000010_predicted_image}} \\
   \includegraphics[align=c,width=\halfimwidth]{\impath{ours/000006_src}} &
   \includegraphics[align=c,width=\imwidth]{\impath{ours/000006_masked_image}} &
   \includegraphics[align=c,width=\imwidth]{\impath{jinx/000006_explicit}} &
   \includegraphics[align=c,width=\imwidth]{\impath{lama/000006_predicted_image}} &
   \includegraphics[align=c,width=\imwidth]{\impath{transformer/000006_predicted_image}} &
   \includegraphics[align=c,width=\imwidth]{\impath{ours/000006_predicted_image}} \\
   \includegraphics[align=c,width=\halfimwidth]{\impath{ours/000011_src}} &
       \includegraphics[align=c,width=\imwidth]{\impath{ours/000011_masked_image}} &
       \includegraphics[align=c,width=\imwidth]{\impath{jinx/000011_explicit}} &
       \includegraphics[align=c,width=\imwidth]{\impath{lama/000011_predicted_image}} &
\includegraphics[align=c,width=\imwidth]{\impath{transformer/000011_predicted_image}} &
       \includegraphics[align=c,width=\imwidth]{\impath{ours/000011_predicted_image}} \\
   \includegraphics[align=c,width=\halfimwidth]{\impath{ours/000030_src}} &
       \includegraphics[align=c,width=\imwidth]{\impath{ours/000030_masked_image}} &
       \includegraphics[align=c,width=\imwidth]{\impath{jinx/000030_explicit}} &
       \includegraphics[align=c,width=\imwidth]{\impath{lama/000030_predicted_image}} &
\includegraphics[align=c,width=\imwidth]{\impath{transformer/000030_predicted_image}} &
       \includegraphics[align=c,width=\imwidth]{\impath{ours/000030_predicted_image}} \\
	\bottomrule
  \end{tabular}\vspace{-1.5em}
\caption{\label{fig:suppinpaintingsamples} Qualitative results on guided image
  inpainting.}
\end{figure*}
}

\newcommand{\figqualinpaintsuppalt}{
  \begin{figure*}[b]
\centering
\scriptsize
	\renewcommand{\imwidth}{0.18\textwidth}
	\renewcommand{\halfimwidth}{0.09\textwidth}
  \renewcommand{\impath}[1]{img/##1} 
	\setlength{\tabcolsep}{0pt}
	\begin{tabular}{c@{\hskip 2pt}c@{\hskip 2pt}c@{\hskip 2pt}ccc}
	\toprule
    keyframes & target & Aligned & LaMa & Transformer & Ours \\
	\midrule
   \includegraphics[align=c,width=\halfimwidth]{\impath{ours/000015_src}} &
       \includegraphics[align=c,width=\imwidth]{\impath{ours/000015_masked_image}} &
       \includegraphics[align=c,width=\imwidth]{\impath{jinx/000015_explicit}} &
       \includegraphics[align=c,width=\imwidth]{\impath{lama/000015_predicted_image}} &
\includegraphics[align=c,width=\imwidth]{\impath{transformer/000015_predicted_image}} &
       \includegraphics[align=c,width=\imwidth]{\impath{ours/000015_predicted_image}} \\
   \includegraphics[align=c,width=\halfimwidth]{\impath{ours/000016_src}} &
       \includegraphics[align=c,width=\imwidth]{\impath{ours/000016_masked_image}} &
       \includegraphics[align=c,width=\imwidth]{\impath{jinx/000016_explicit}} &
       \includegraphics[align=c,width=\imwidth]{\impath{lama/000016_predicted_image}} &
\includegraphics[align=c,width=\imwidth]{\impath{transformer/000016_predicted_image}} &
       \includegraphics[align=c,width=\imwidth]{\impath{ours/000016_predicted_image}} \\
   \includegraphics[align=c,width=\halfimwidth]{\impath{ours/000017_src}} &
       \includegraphics[align=c,width=\imwidth]{\impath{ours/000017_masked_image}} &
       \includegraphics[align=c,width=\imwidth]{\impath{jinx/000017_explicit}} &
       \includegraphics[align=c,width=\imwidth]{\impath{lama/000017_predicted_image}} &
\includegraphics[align=c,width=\imwidth]{\impath{transformer/000017_predicted_image}} &
       \includegraphics[align=c,width=\imwidth]{\impath{ours/000017_predicted_image}} \\
   \includegraphics[align=c,width=\halfimwidth]{\impath{ours/000018_src}} &
       \includegraphics[align=c,width=\imwidth]{\impath{ours/000018_masked_image}} &
       \includegraphics[align=c,width=\imwidth]{\impath{jinx/000018_explicit}} &
       \includegraphics[align=c,width=\imwidth]{\impath{lama/000018_predicted_image}} &
\includegraphics[align=c,width=\imwidth]{\impath{transformer/000018_predicted_image}} &
       \includegraphics[align=c,width=\imwidth]{\impath{ours/000018_predicted_image}} \\
   \includegraphics[align=c,width=\halfimwidth]{\impath{ours/000044_src}} &
       \includegraphics[align=c,width=\imwidth]{\impath{ours/000044_masked_image}} &
       \includegraphics[align=c,width=\imwidth]{\impath{jinx/000044_explicit}} &
       \includegraphics[align=c,width=\imwidth]{\impath{lama/000044_predicted_image}} &
\includegraphics[align=c,width=\imwidth]{\impath{transformer/000044_predicted_image}} &
       \includegraphics[align=c,width=\imwidth]{\impath{ours/000044_predicted_image}} \\
   \includegraphics[align=c,width=\halfimwidth]{\impath{ours/000047_src}} &
       \includegraphics[align=c,width=\imwidth]{\impath{ours/000047_masked_image}} &
       \includegraphics[align=c,width=\imwidth]{\impath{jinx/000047_explicit}} &
       \includegraphics[align=c,width=\imwidth]{\impath{lama/000047_predicted_image}} &
\includegraphics[align=c,width=\imwidth]{\impath{transformer/000047_predicted_image}} &
       \includegraphics[align=c,width=\imwidth]{\impath{ours/000047_predicted_image}} \\
	\bottomrule
  \end{tabular}\vspace{-1.5em}
\caption{\label{fig:suppinpaintingsamplesalt} Qualitative results on guided image
  inpainting.}
\end{figure*}
}

\newcommand{\suppfigqualinpaint}{
\begin{figure}
\centering
\scriptsize
	\renewcommand{\imwidth}{0.19\textwidth}
	\renewcommand{\halfimwidth}{0.095\textwidth}
  \renewcommand{\impath}[1]{img/##1} 
	\setlength{\tabcolsep}{0pt}
	\begin{tabular}{c@{\hskip 2pt}c@{\hskip 2pt}cccc}
	\toprule
    keyframes & target & Aligned & LaMa & Transformer & Ours \\
	\midrule
   \includegraphics[align=c,width=\halfimwidth]{\impath{ours/000000_src}} &
   \includegraphics[align=c,width=\imwidth]{\impath{ours/000000_masked_image}} &
   \includegraphics[align=c,width=\imwidth]{\impath{jinx/000000_explicit}} &
   \includegraphics[align=c,width=\imwidth]{\impath{lama/000000_predicted_image}} &
   \includegraphics[align=c,width=\imwidth]{\impath{transformer/000000_predicted_image}} &
   \includegraphics[align=c,width=\imwidth]{\impath{ours/000000_predicted_image}} \\
   \includegraphics[align=c,width=\halfimwidth]{\impath{ours/000002_src}} &
   \includegraphics[align=c,width=\imwidth]{\impath{ours/000002_masked_image}} &
   \includegraphics[align=c,width=\imwidth]{\impath{jinx/000002_explicit}} &
   \includegraphics[align=c,width=\imwidth]{\impath{lama/000002_predicted_image}} &
   \includegraphics[align=c,width=\imwidth]{\impath{transformer/000002_predicted_image}} &
   \includegraphics[align=c,width=\imwidth]{\impath{ours/000002_predicted_image}} \\
   \includegraphics[align=c,width=\halfimwidth]{\impath{ours/000008_src}} &
       \includegraphics[align=c,width=\imwidth]{\impath{ours/000008_masked_image}} &
       \includegraphics[align=c,width=\imwidth]{\impath{jinx/000008_explicit}} &
       \includegraphics[align=c,width=\imwidth]{\impath{lama/000008_predicted_image}} &
\includegraphics[align=c,width=\imwidth]{\impath{transformer/000008_predicted_image}} &
       \includegraphics[align=c,width=\imwidth]{\impath{ours/000008_predicted_image}} \\
   \includegraphics[align=c,width=\halfimwidth]{\impath{ours/000013_src}} &
       \includegraphics[align=c,width=\imwidth]{\impath{ours/000013_masked_image}} &
       \includegraphics[align=c,width=\imwidth]{\impath{jinx/000013_explicit}} &
       \includegraphics[align=c,width=\imwidth]{\impath{lama/000013_predicted_image}} &
\includegraphics[align=c,width=\imwidth]{\impath{transformer/000013_predicted_image}} &
       \includegraphics[align=c,width=\imwidth]{\impath{ours/000013_predicted_image}} \\
   \includegraphics[align=c,width=\halfimwidth]{\impath{ours/000036_src}} &
       \includegraphics[align=c,width=\imwidth]{\impath{ours/000036_masked_image}} &
       \includegraphics[align=c,width=\imwidth]{\impath{jinx/000036_explicit}} &
       \includegraphics[align=c,width=\imwidth]{\impath{lama/000036_predicted_image}} &
\includegraphics[align=c,width=\imwidth]{\impath{transformer/000036_predicted_image}} &
       \includegraphics[align=c,width=\imwidth]{\impath{ours/000036_predicted_image}} \\
	\bottomrule
  \end{tabular}\vspace{-1.25em}
\caption{\label{fig:suppfigqualinpaint} A larger version of
Fig.~\ref{figqualinpaint}.\vspace{-1.5em}}
\end{figure}
}

\newcommand{\figqualitativevibear}{
  \begin{figure*}
    \centering
	\renewcommand{\imwidth}{0.23\textwidth}
	\renewcommand{\davisseq}{davis_bear}
	\setlength{\tabcolsep}{0pt}
	\begin{tabular}{ccccc}
	\toprule
     & \multicolumn{4}{c}{Frames}\\
     \midrule
input &
\includegraphics[align=c,width=\imwidth]{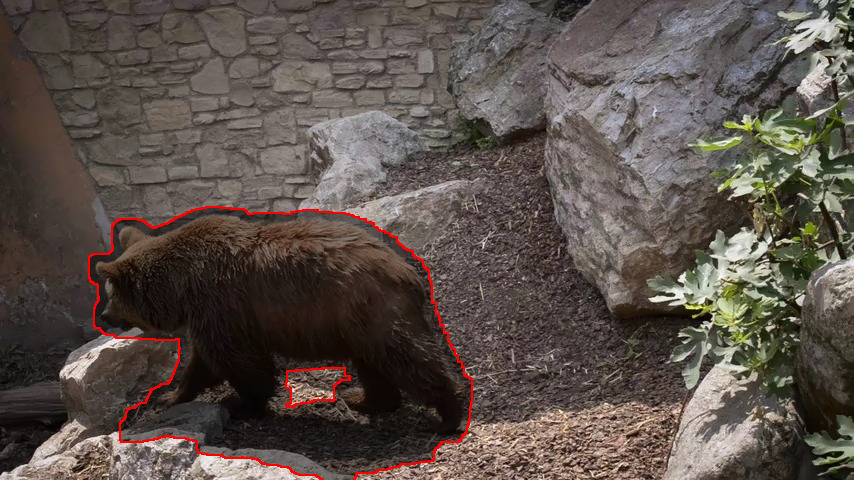} &
\includegraphics[align=c,width=\imwidth]{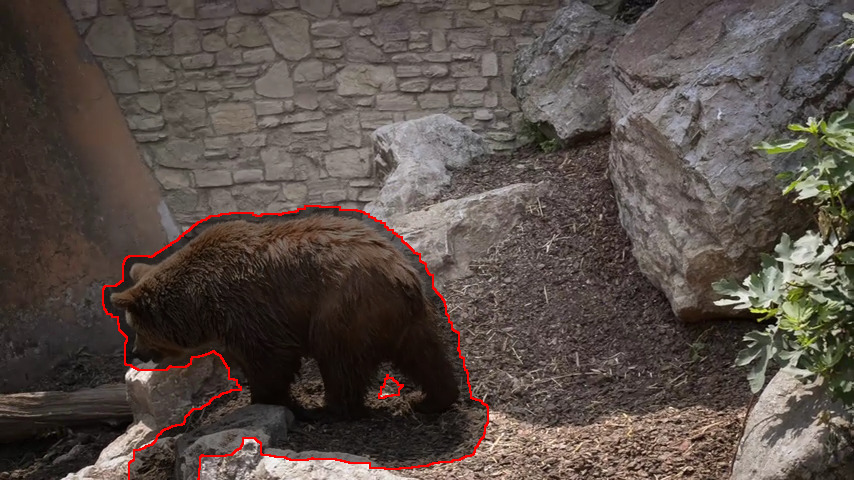} &
\includegraphics[align=c,width=\imwidth]{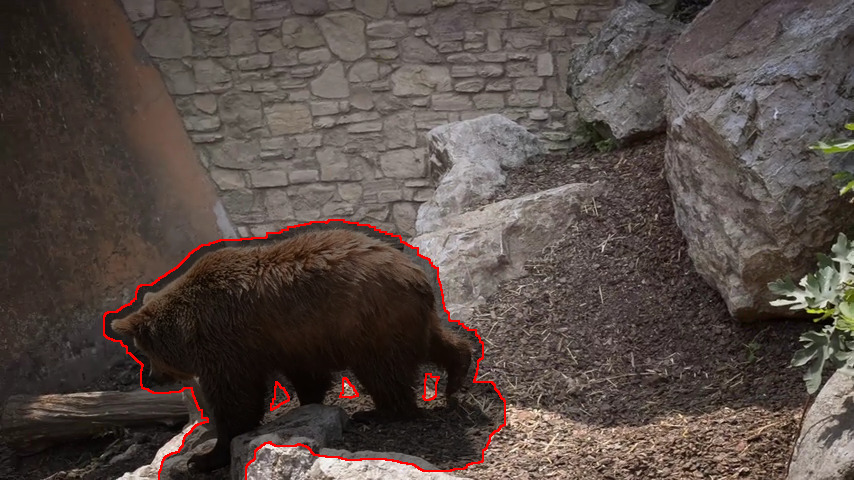} &
\includegraphics[align=c,width=\imwidth]{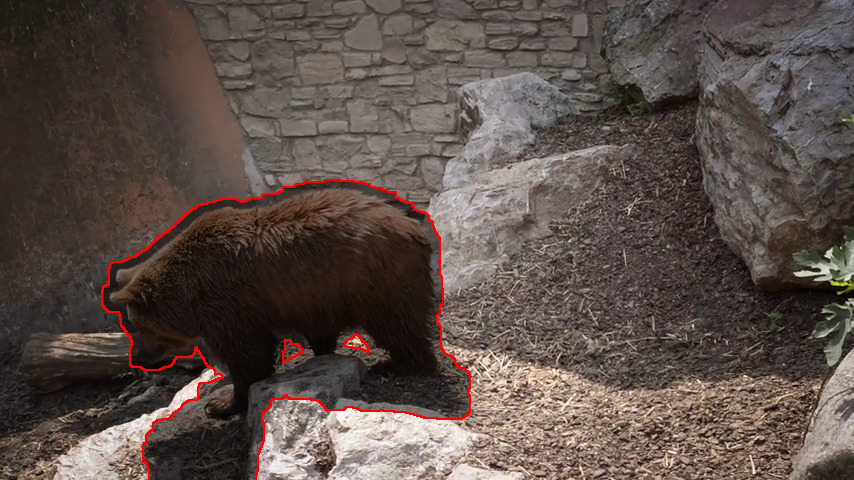} \\
Ours \scriptsize{14+prealign} &
\includegraphics[align=c,width=\imwidth]{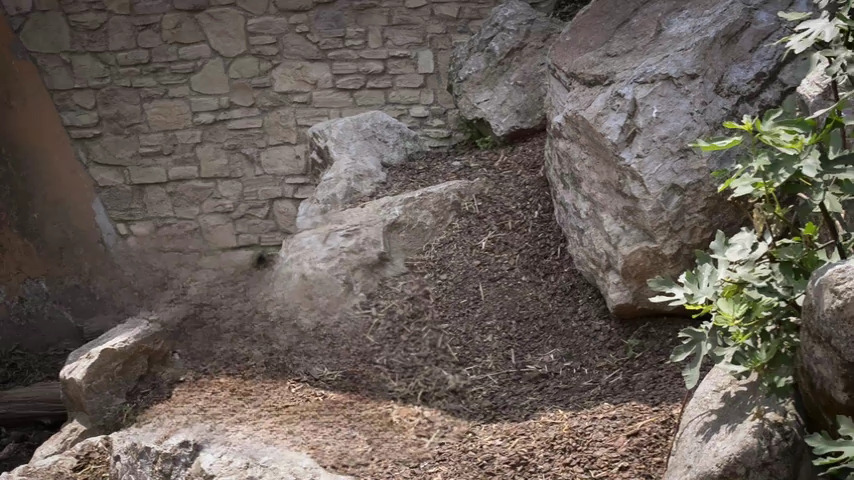} &
\includegraphics[align=c,width=\imwidth]{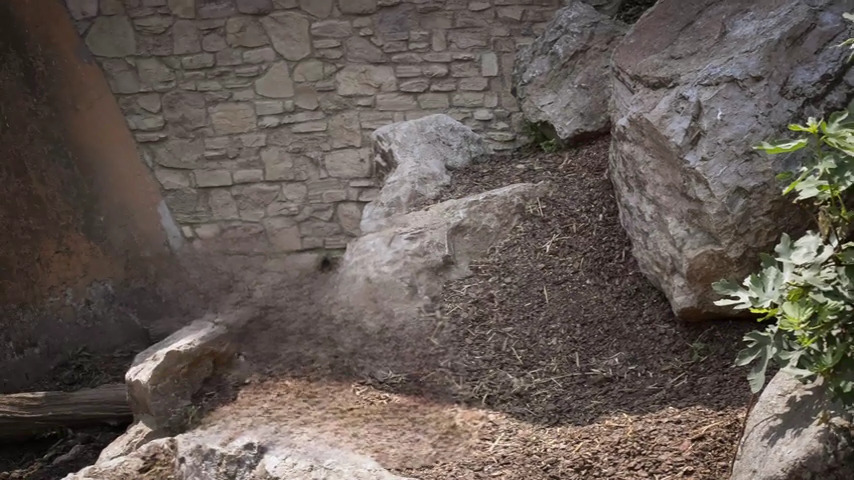} &
\includegraphics[align=c,width=\imwidth]{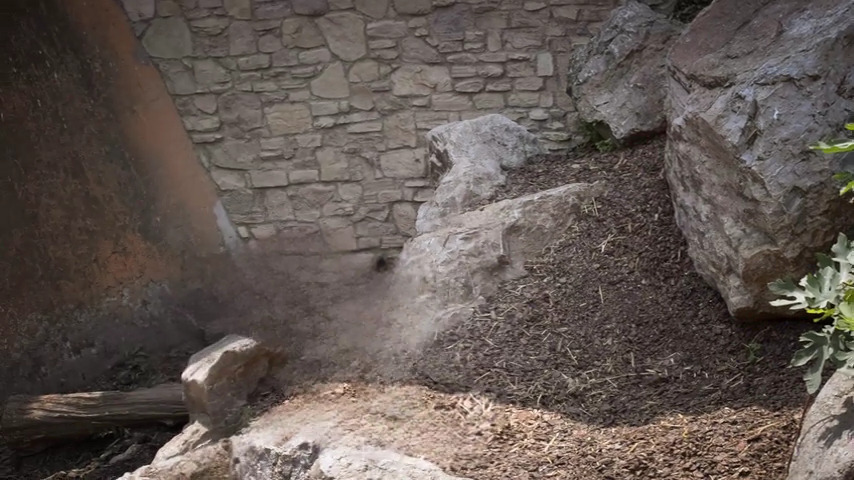} &
\includegraphics[align=c,width=\imwidth]{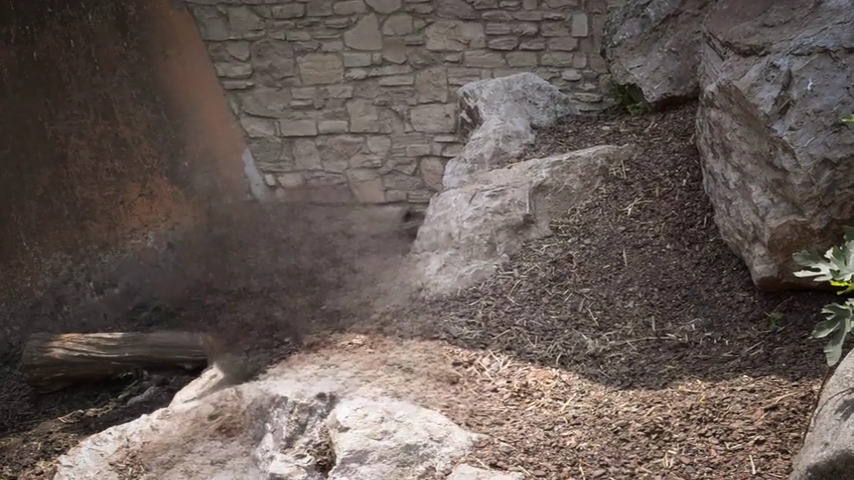} \\
Ours \scriptsize{14} &
\includegraphics[align=c,width=\imwidth]{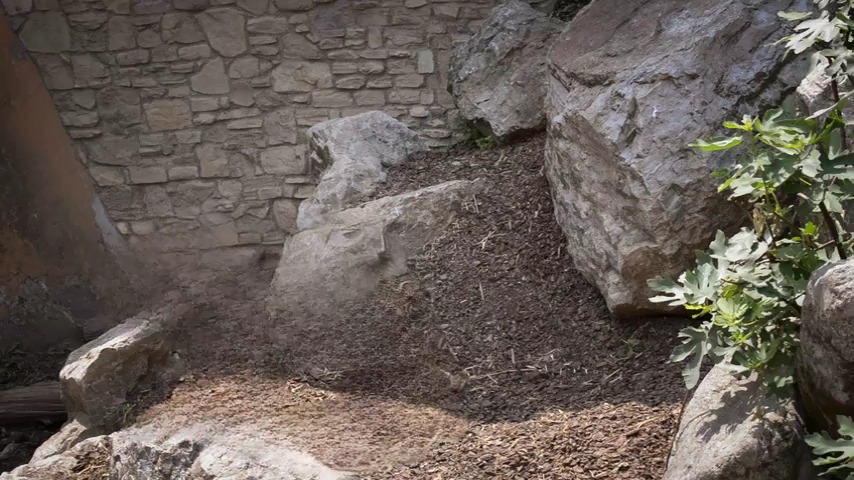} &
\includegraphics[align=c,width=\imwidth]{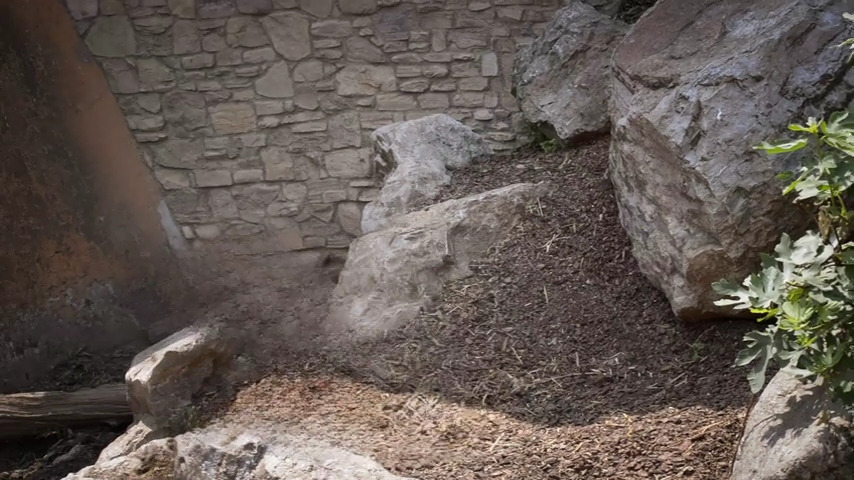} &
\includegraphics[align=c,width=\imwidth]{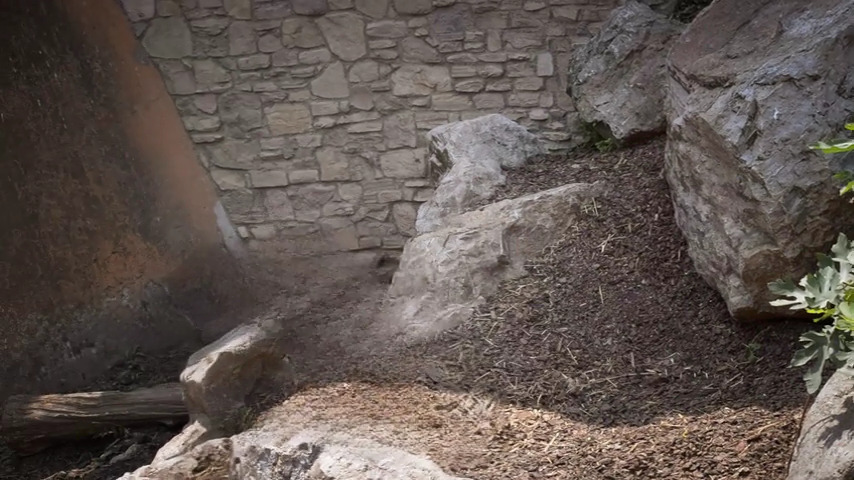} &
\includegraphics[align=c,width=\imwidth]{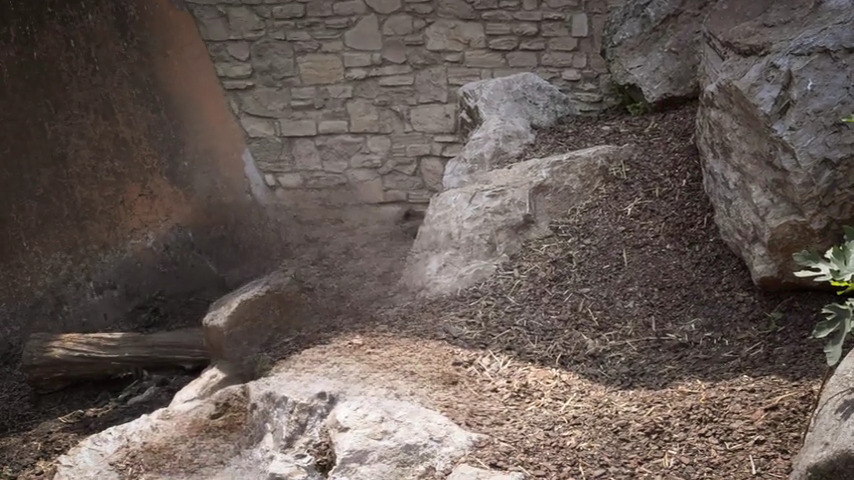} \\
LaMa\cite{lama} \scriptsize{+flow} &
\includegraphics[align=c,width=\imwidth]{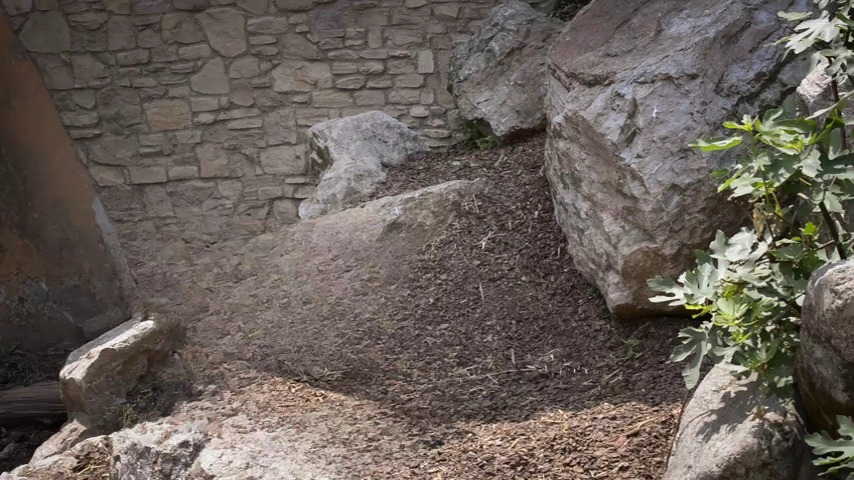} &
\includegraphics[align=c,width=\imwidth]{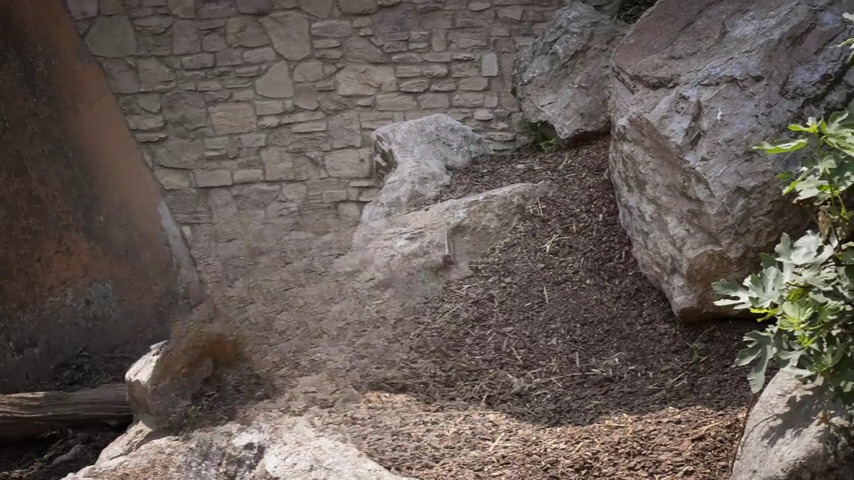} &
\includegraphics[align=c,width=\imwidth]{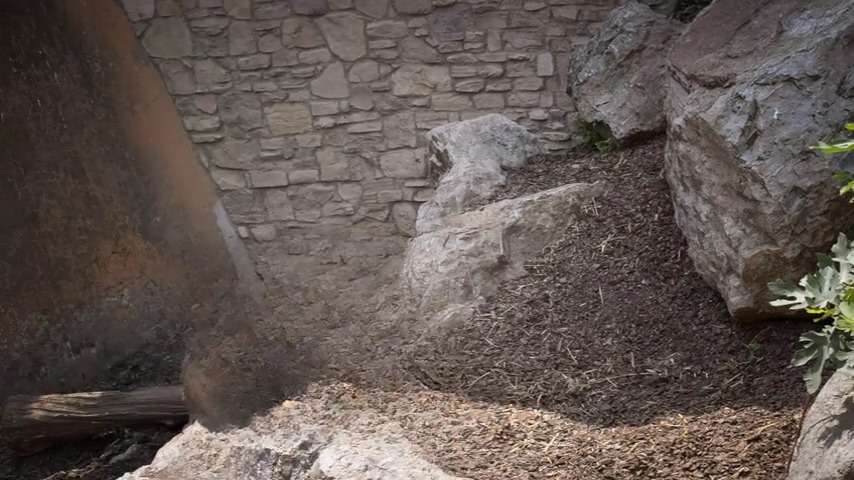} &
\includegraphics[align=c,width=\imwidth]{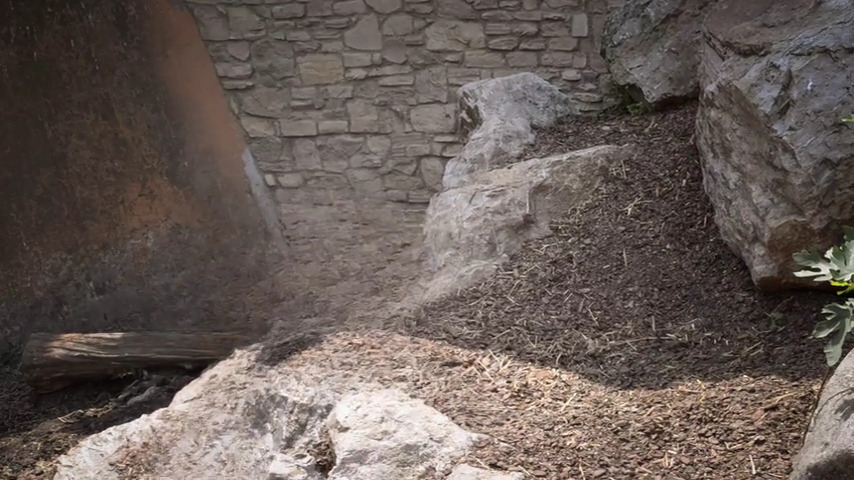} \\
DFCNet\cite{dfcnet} &
\includegraphics[align=c,width=\imwidth]{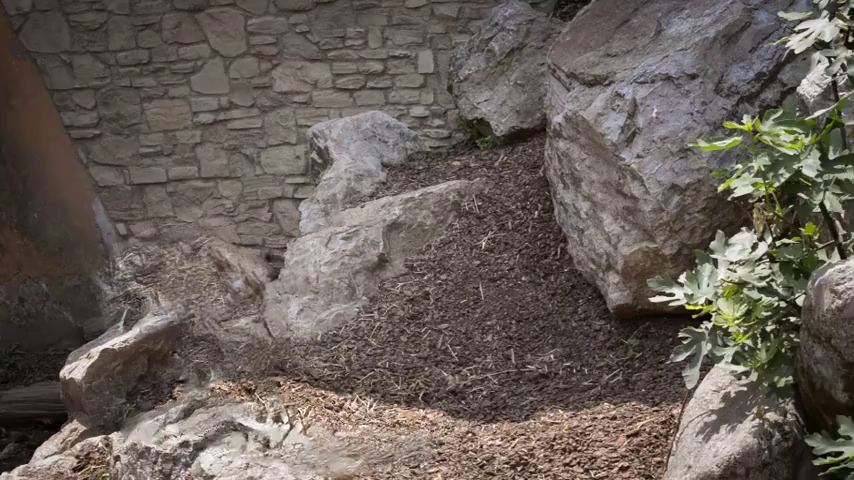} &
\includegraphics[align=c,width=\imwidth]{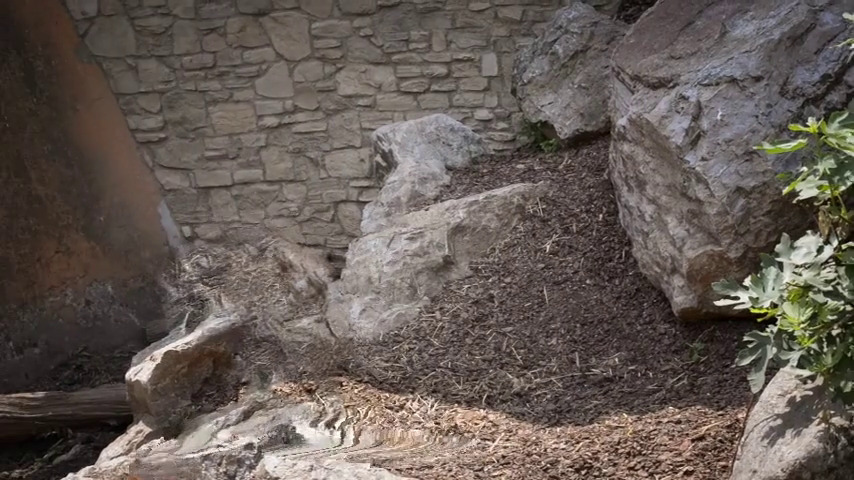} &
\includegraphics[align=c,width=\imwidth]{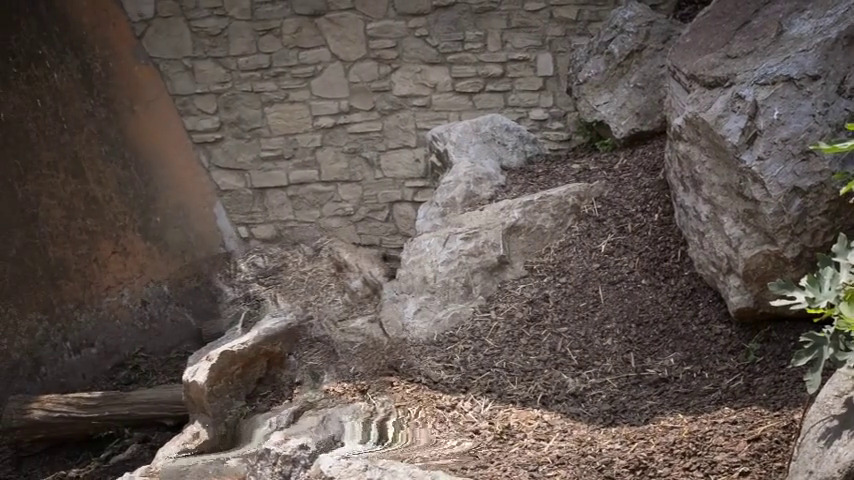} &
\includegraphics[align=c,width=\imwidth]{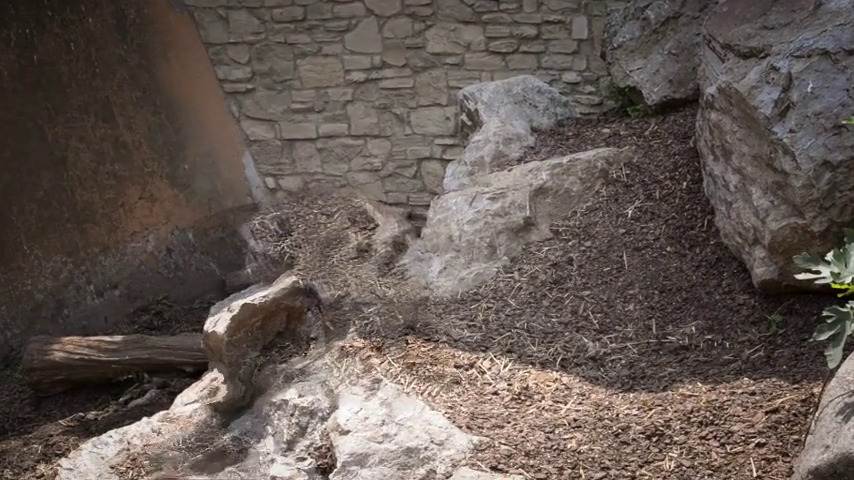} \\
E2FGVI\cite{e2fgvi} &
\includegraphics[align=c,width=\imwidth]{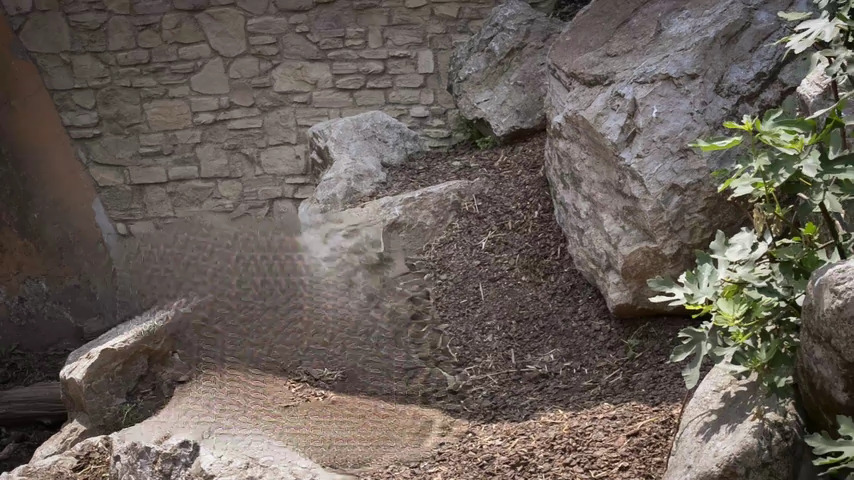} &
\includegraphics[align=c,width=\imwidth]{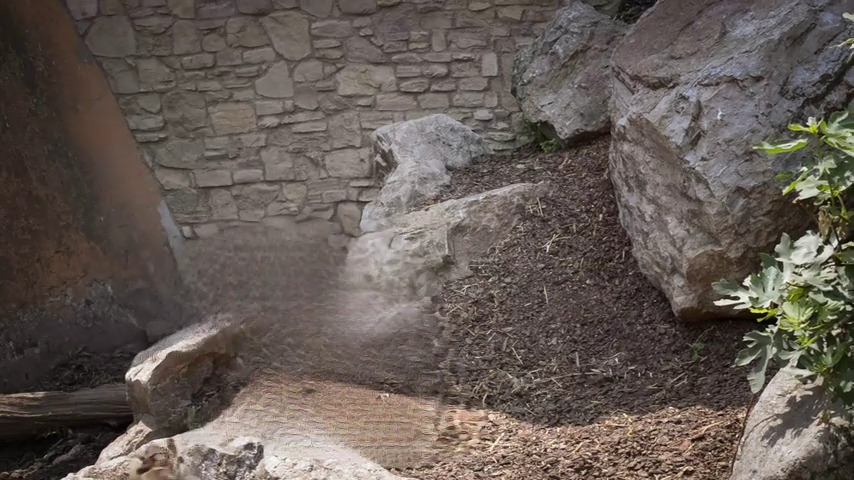} &
\includegraphics[align=c,width=\imwidth]{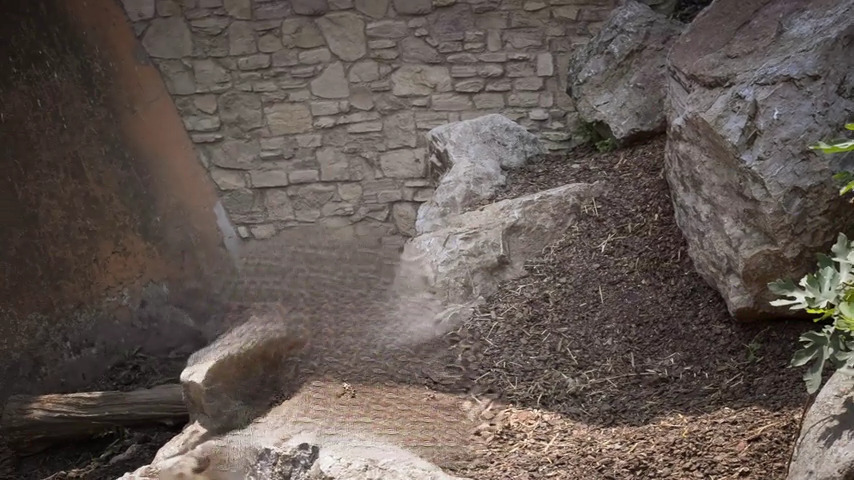} &
\includegraphics[align=c,width=\imwidth]{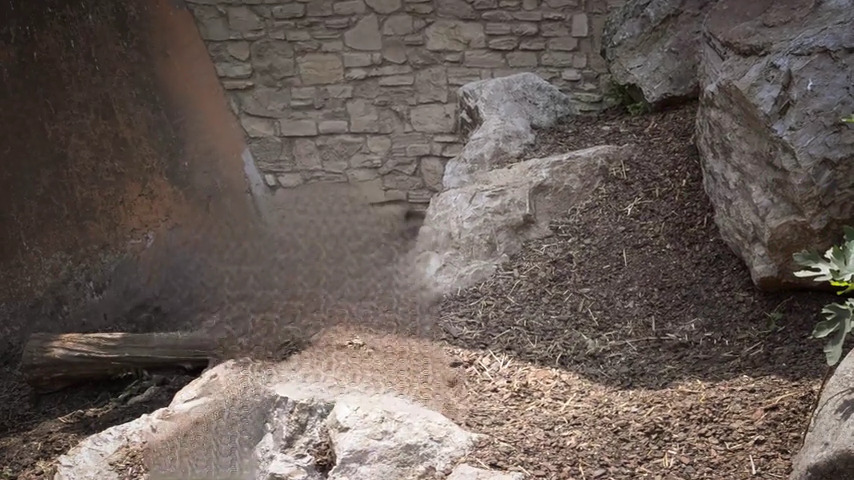} \\
FGVC\cite{fgvc} &
\includegraphics[align=c,width=\imwidth]{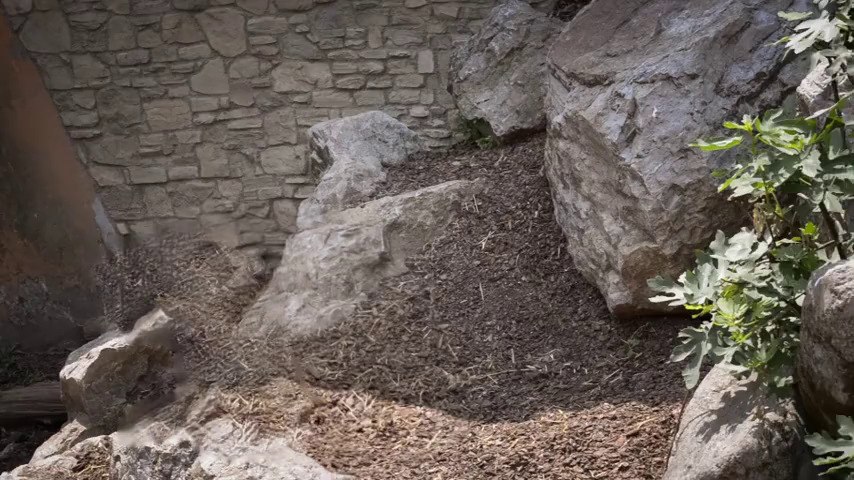} &
\includegraphics[align=c,width=\imwidth]{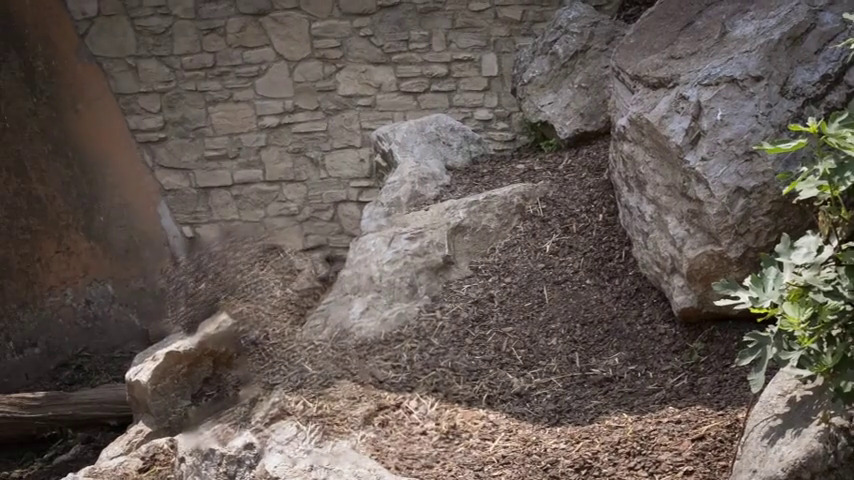} &
\includegraphics[align=c,width=\imwidth]{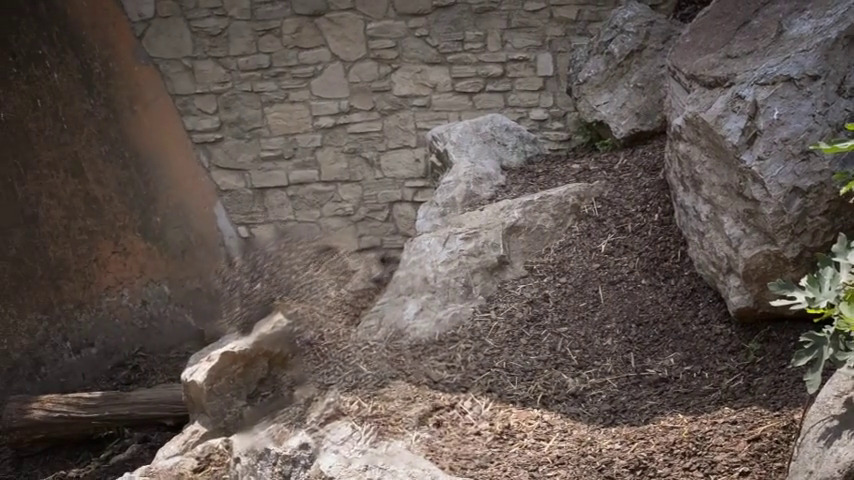} &
\includegraphics[align=c,width=\imwidth]{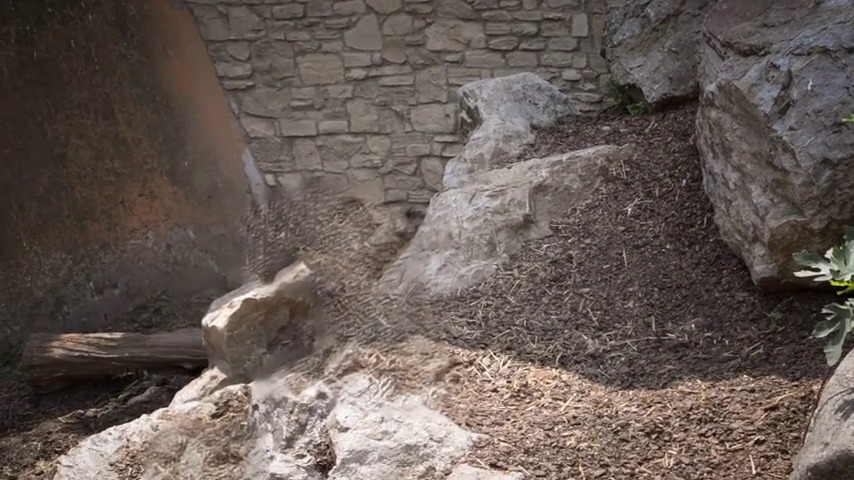} \\
	\bottomrule
  \end{tabular}
\caption{\label{fig:qualitativevibear} Qualitative results for video inpainting on
    the DAVIS\cite{davis} sequence \emph{bear}. Animated version at \url{\projecturl}.}
\end{figure*}
}

\newcommand{\figqualitativevibdf}{
  \begin{figure*}
    \centering
	\renewcommand{\imwidth}{0.23\textwidth}
	\renewcommand{\davisseq}{davis_breakdance-flare}
	\setlength{\tabcolsep}{0pt}
	\begin{tabular}{ccccc}
	\toprule
     & \multicolumn{4}{c}{Frames}\\
     \midrule
input &
\includegraphics[align=c,width=\imwidth]{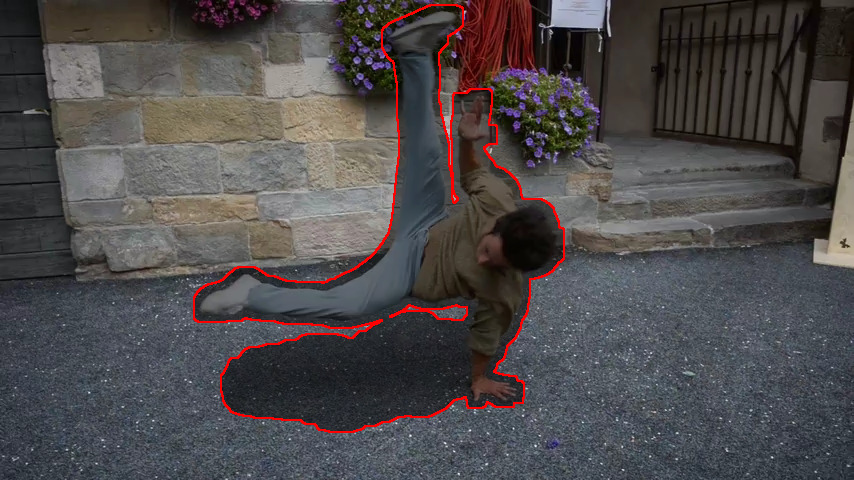} &
\includegraphics[align=c,width=\imwidth]{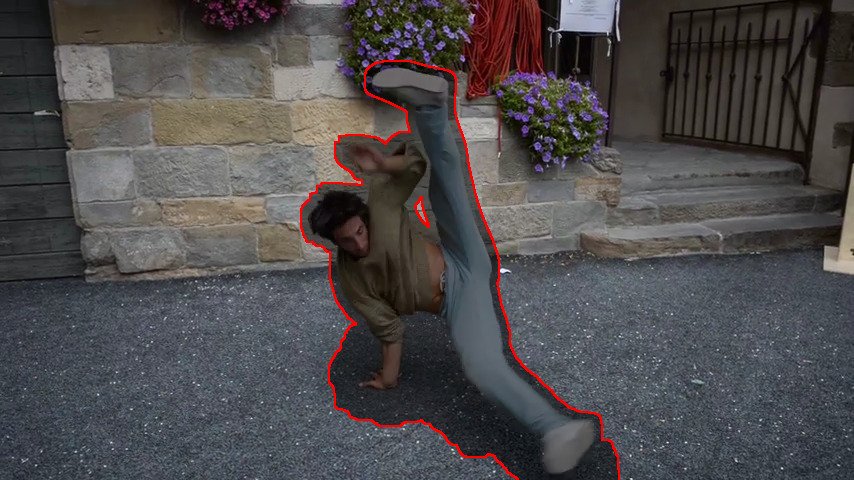} &
\includegraphics[align=c,width=\imwidth]{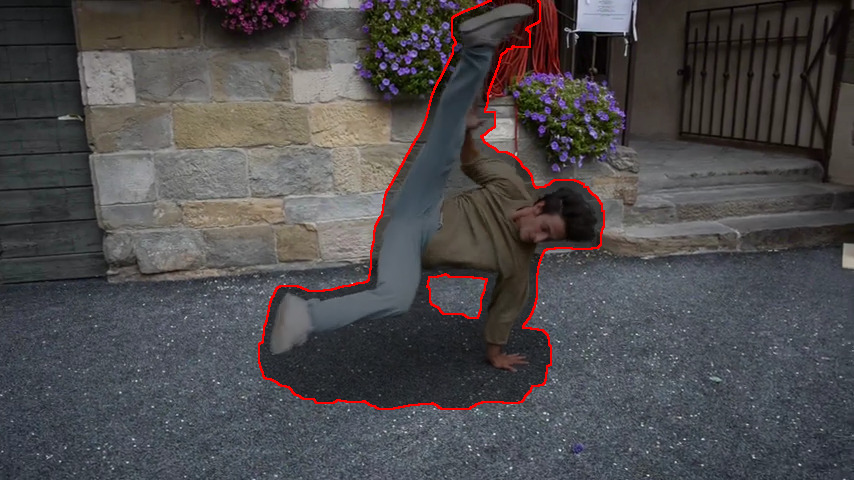} &
\includegraphics[align=c,width=\imwidth]{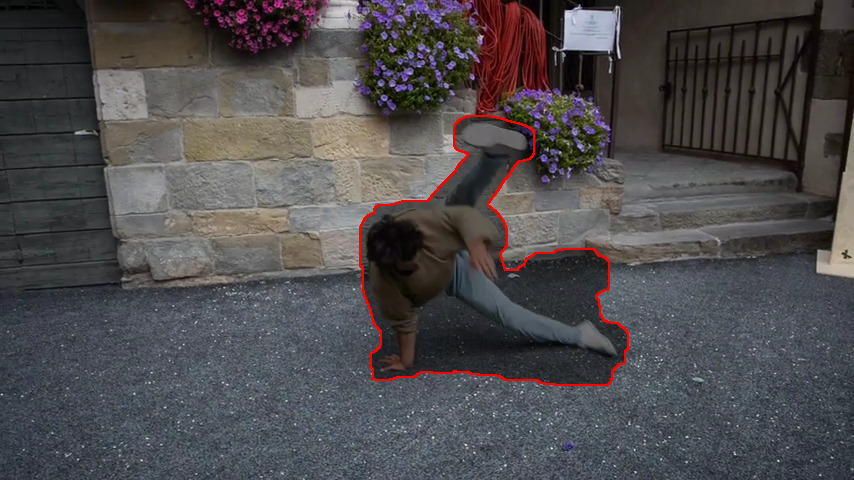} \\
Ours \scriptsize{14+prealign} &
\includegraphics[align=c,width=\imwidth]{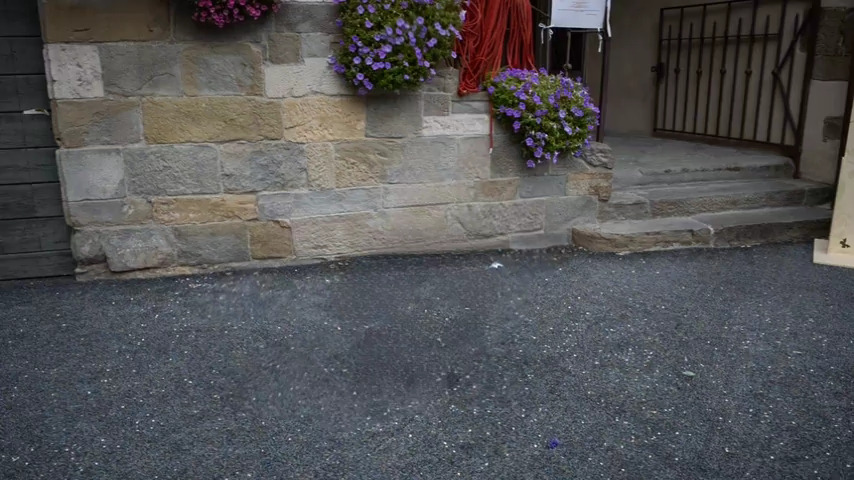} &
\includegraphics[align=c,width=\imwidth]{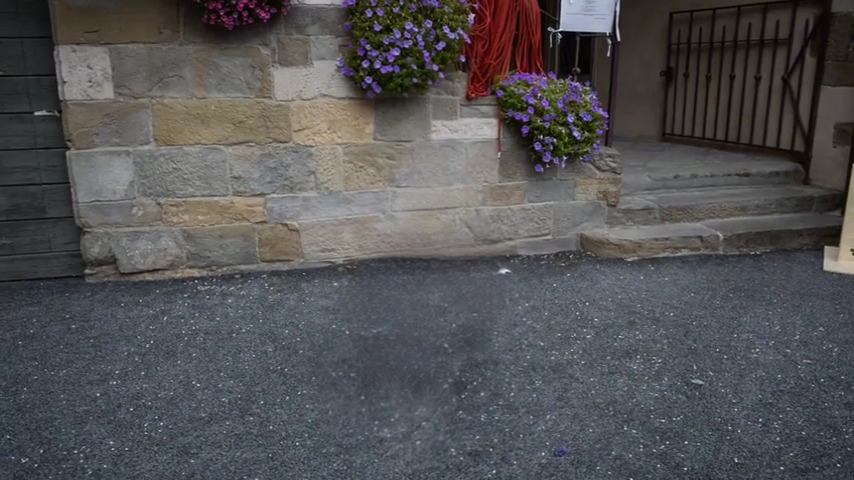} &
\includegraphics[align=c,width=\imwidth]{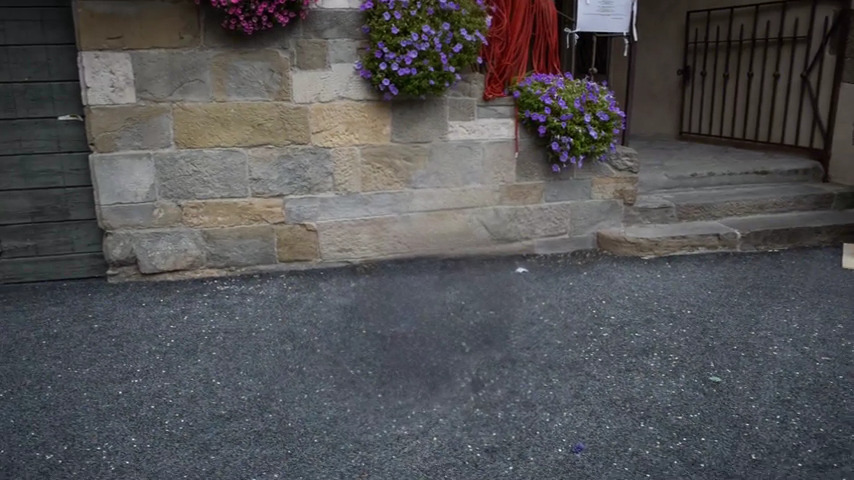} &
\includegraphics[align=c,width=\imwidth]{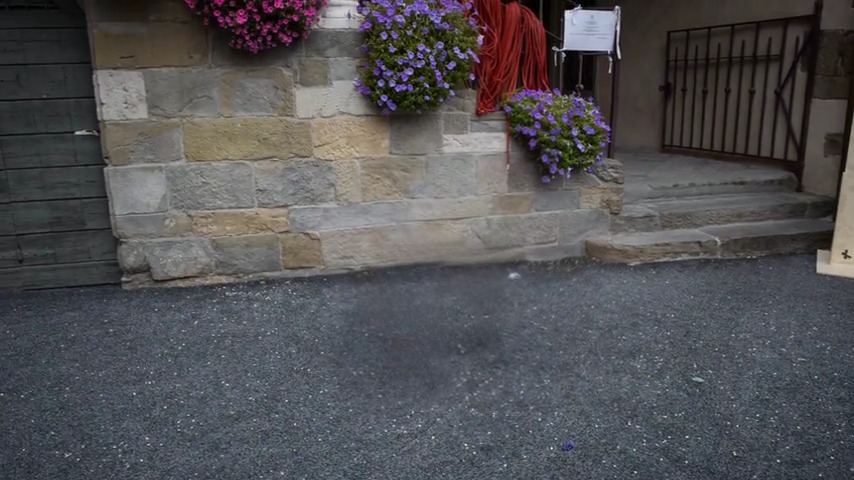} \\
Ours \scriptsize{14} &
\includegraphics[align=c,width=\imwidth]{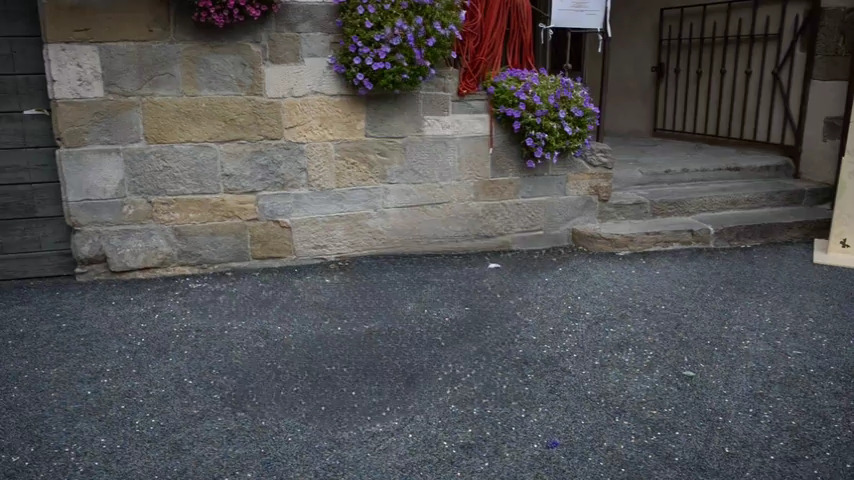} &
\includegraphics[align=c,width=\imwidth]{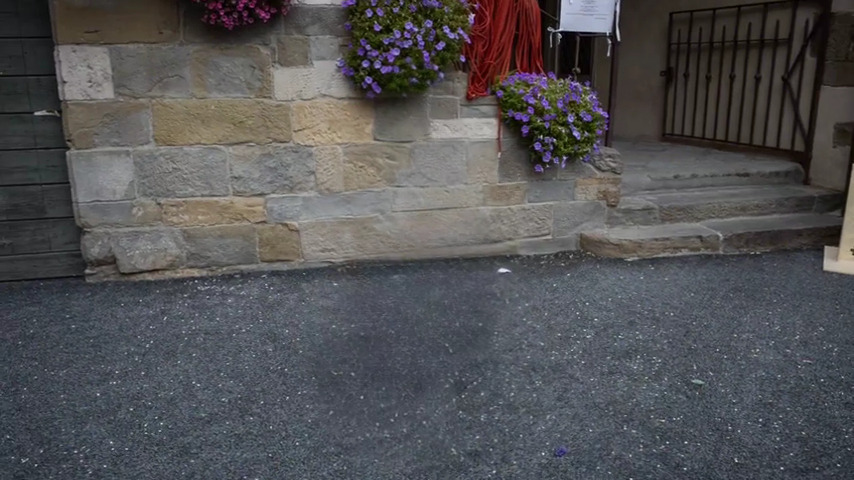} &
\includegraphics[align=c,width=\imwidth]{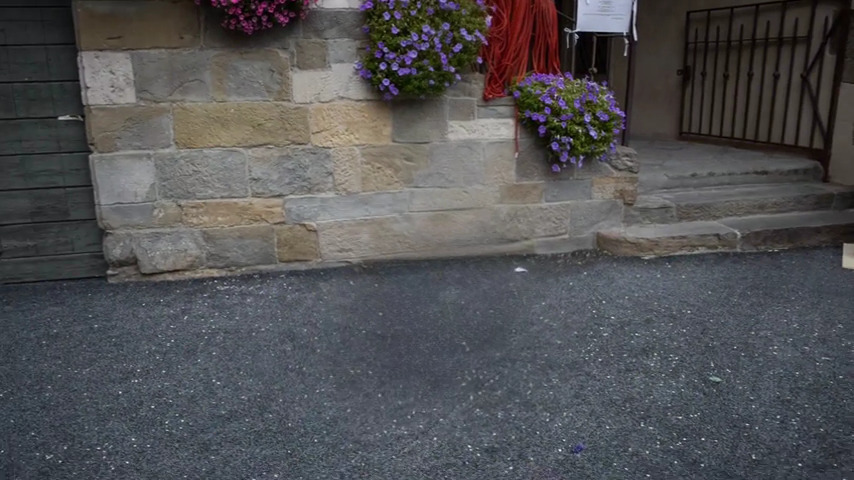} &
\includegraphics[align=c,width=\imwidth]{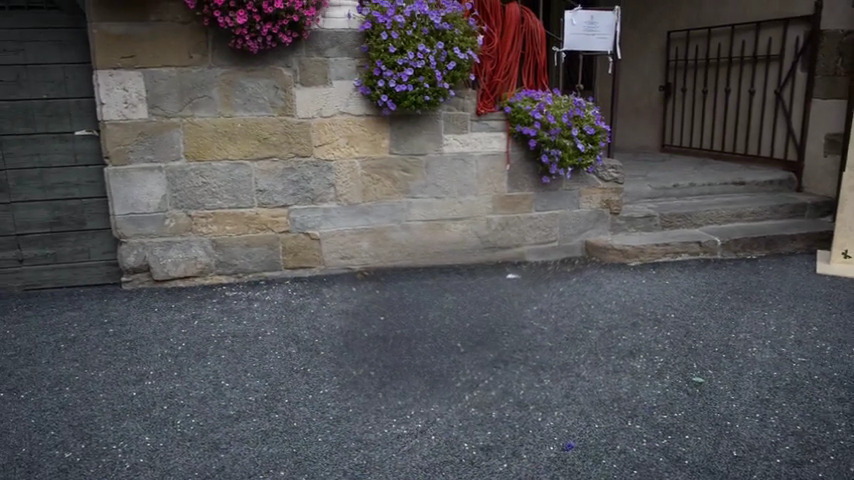} \\
LaMa\cite{lama} \scriptsize{+flow} &
\includegraphics[align=c,width=\imwidth]{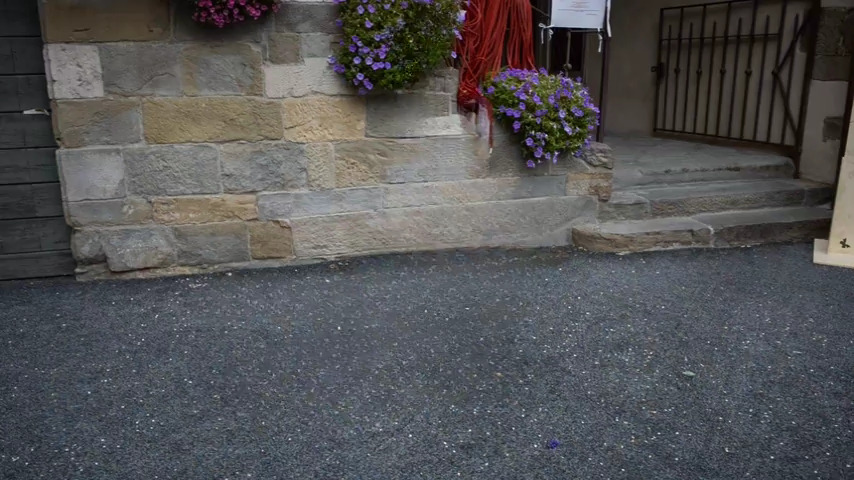} &
\includegraphics[align=c,width=\imwidth]{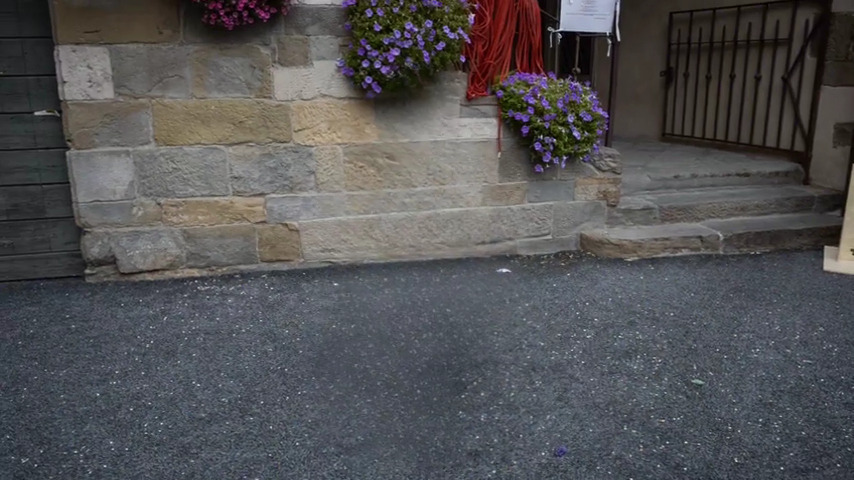} &
\includegraphics[align=c,width=\imwidth]{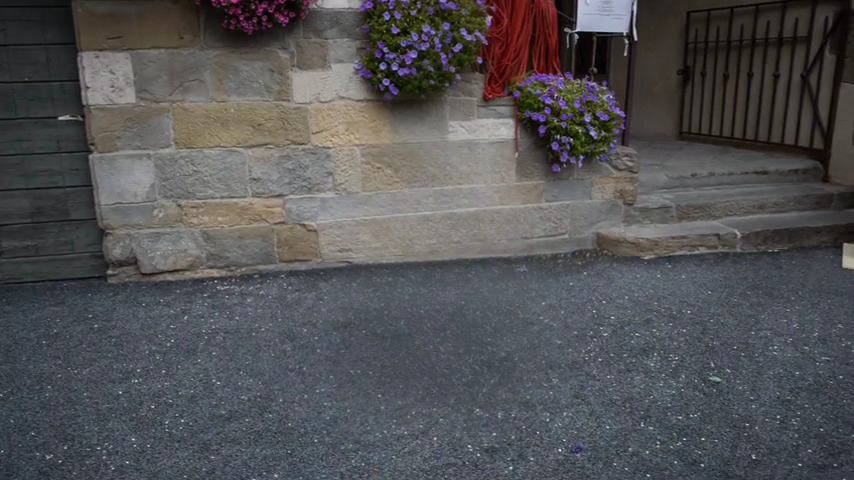} &
\includegraphics[align=c,width=\imwidth]{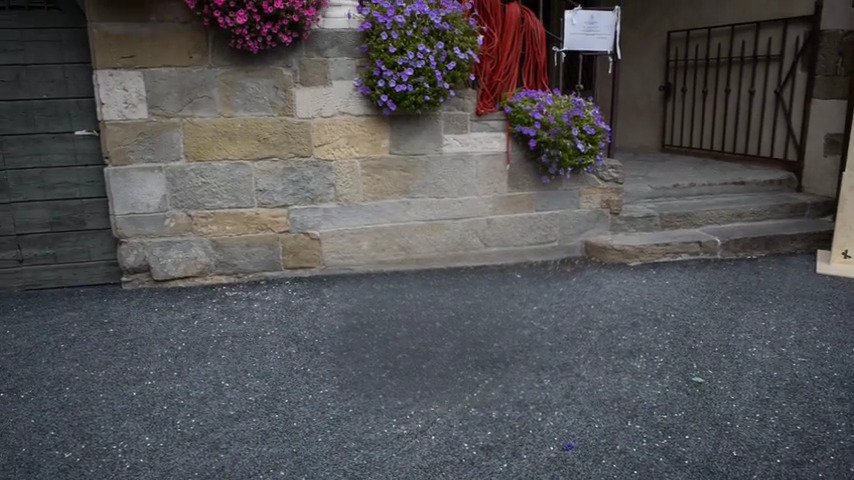} \\
DFCNet\cite{dfcnet} &
\includegraphics[align=c,width=\imwidth]{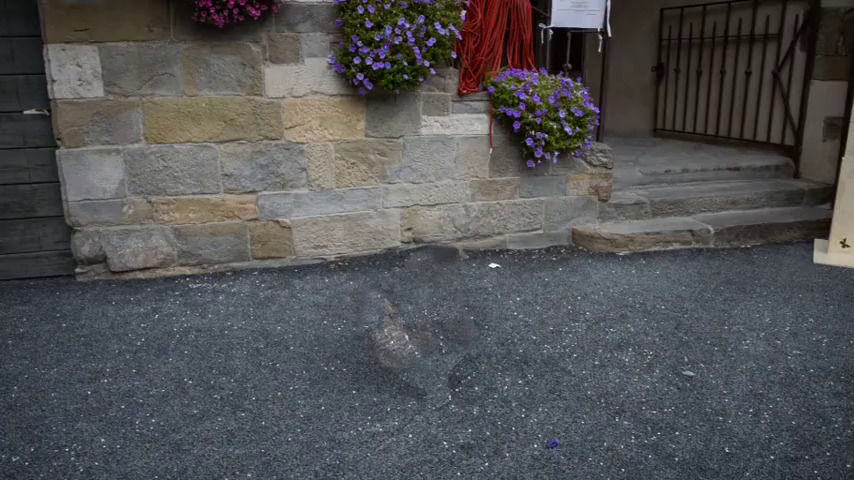} &
\includegraphics[align=c,width=\imwidth]{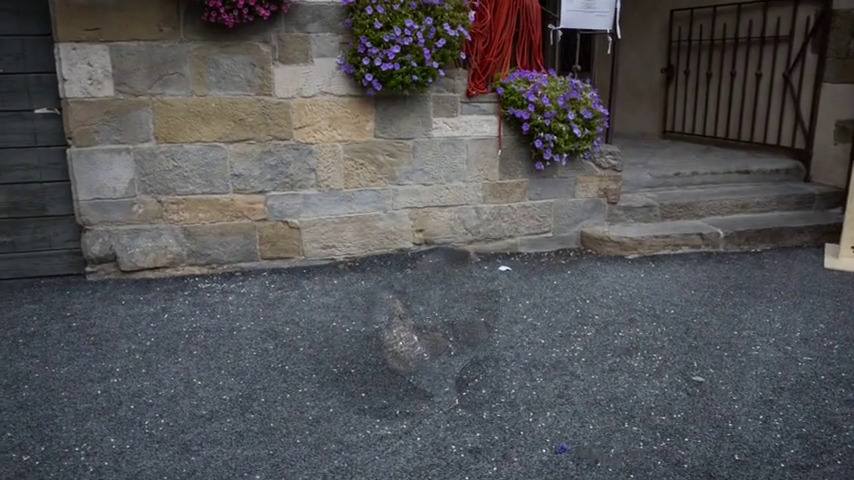} &
\includegraphics[align=c,width=\imwidth]{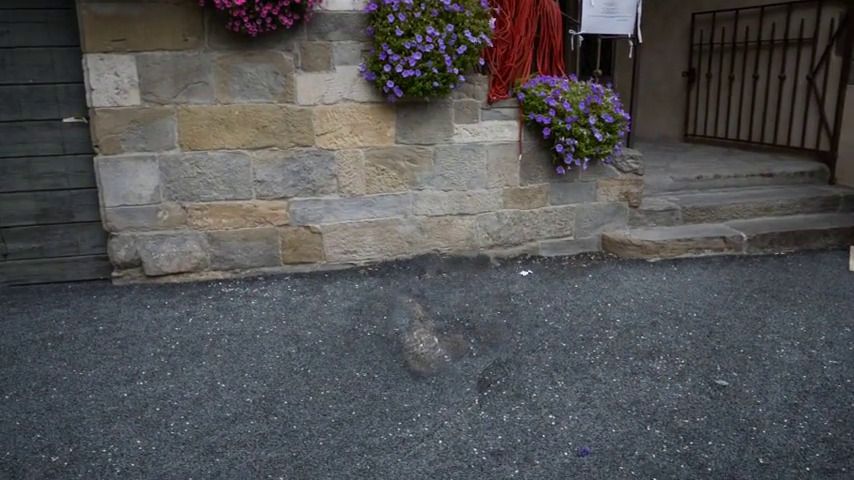} &
\includegraphics[align=c,width=\imwidth]{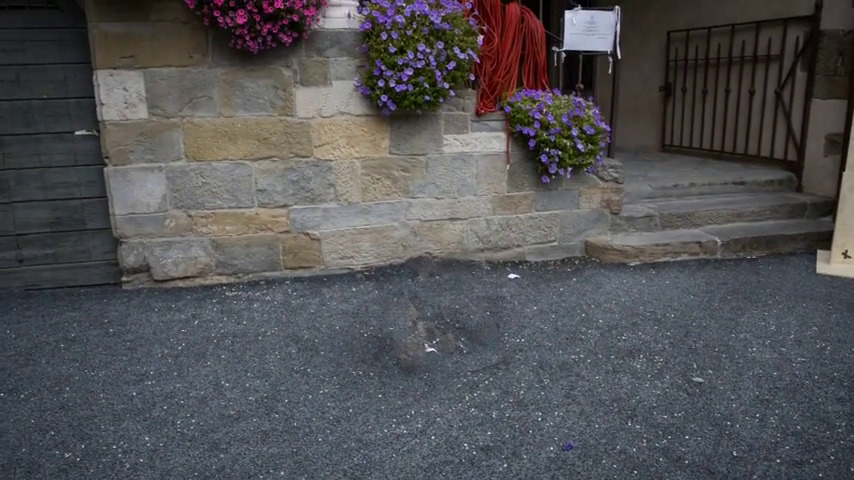} \\
E2FGVI\cite{e2fgvi} &
\includegraphics[align=c,width=\imwidth]{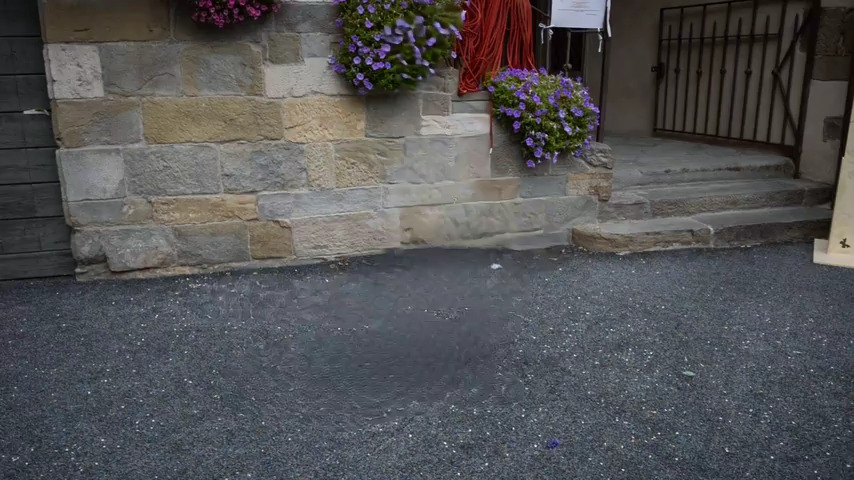} &
\includegraphics[align=c,width=\imwidth]{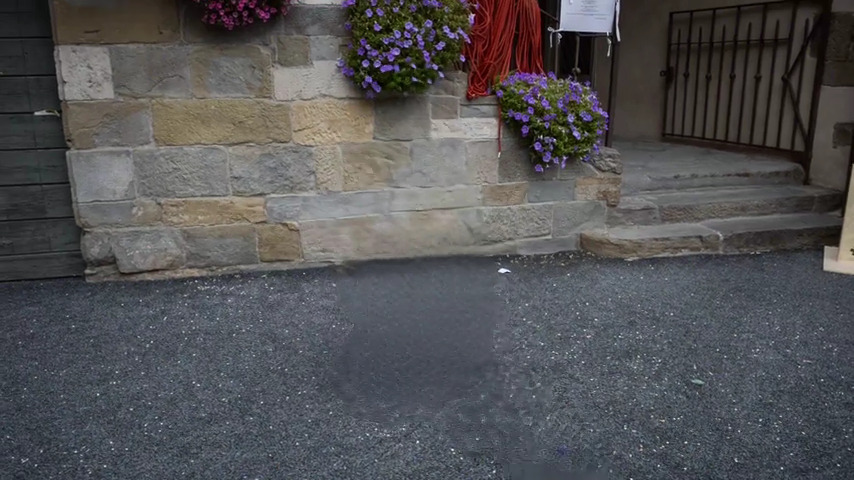} &
\includegraphics[align=c,width=\imwidth]{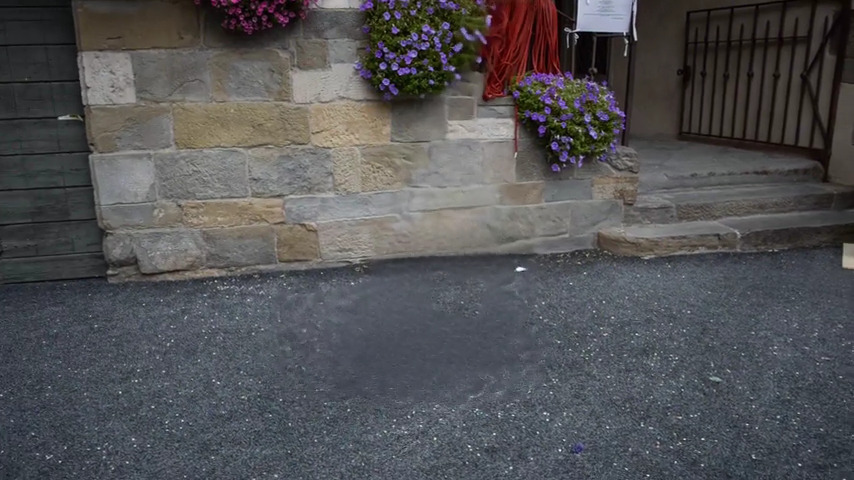} &
\includegraphics[align=c,width=\imwidth]{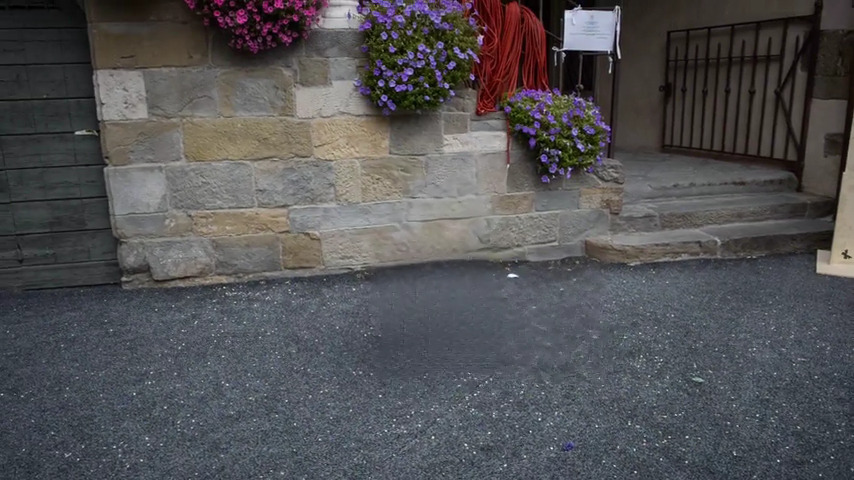} \\
FGVC\cite{fgvc} &
\includegraphics[align=c,width=\imwidth]{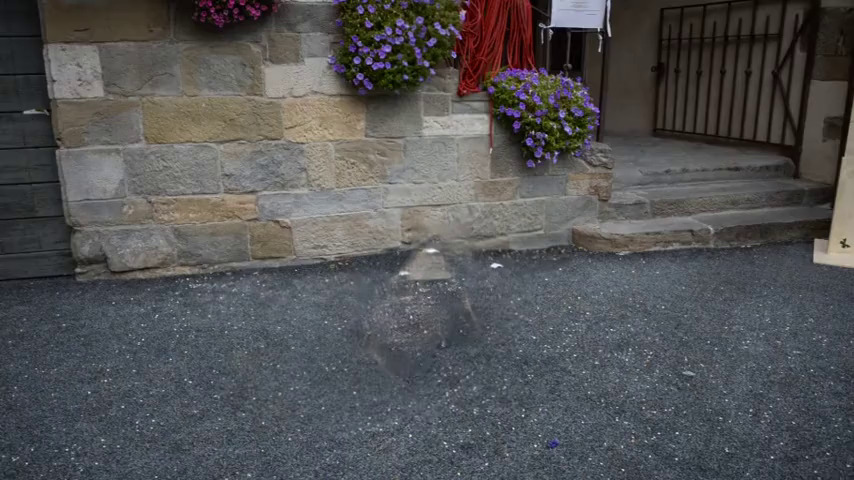} &
\includegraphics[align=c,width=\imwidth]{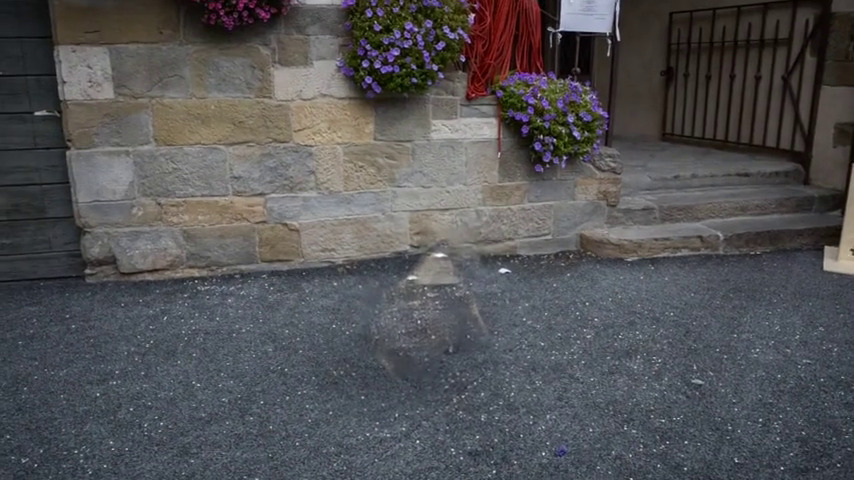} &
\includegraphics[align=c,width=\imwidth]{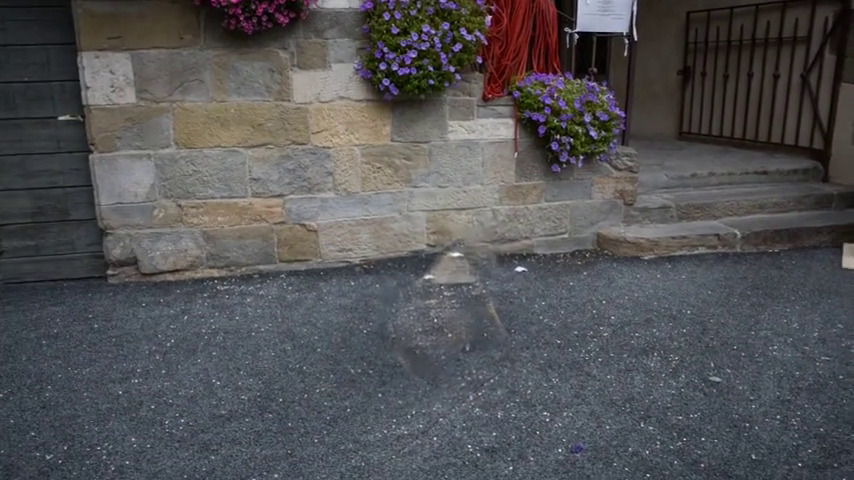} &
\includegraphics[align=c,width=\imwidth]{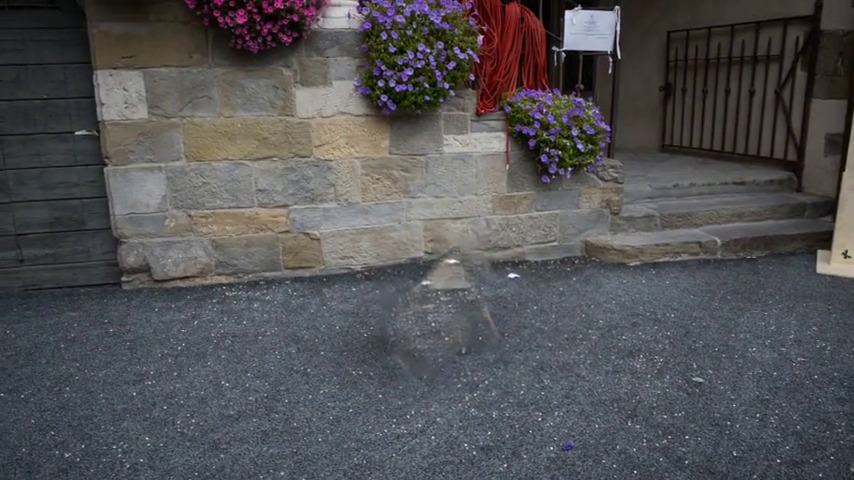} \\
	\bottomrule
  \end{tabular}
\caption{\label{fig:qualitativevibdf} Qualitative results for video inpainting on
    the DAVIS\cite{davis} sequence \emph{breakdance-flare}. Animated version at \url{\projecturl}.}
\end{figure*}
}

\newcommand{\figqualitativevibd}{
  \begin{figure*}
    \centering
	\renewcommand{\imwidth}{0.23\textwidth}
	\renewcommand{\davisseq}{davis_breakdance}
	\setlength{\tabcolsep}{0pt}
	\begin{tabular}{ccccc}
	\toprule
     & \multicolumn{4}{c}{Frames}\\
     \midrule
input &
\includegraphics[align=c,width=\imwidth]{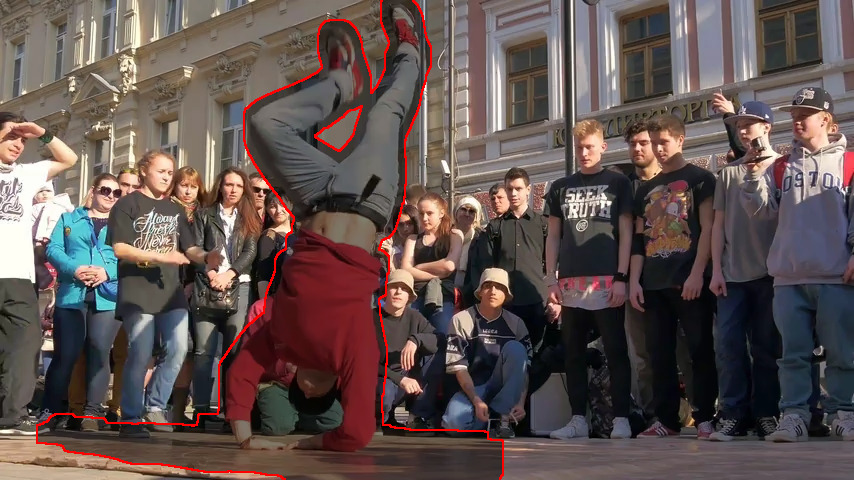} &
\includegraphics[align=c,width=\imwidth]{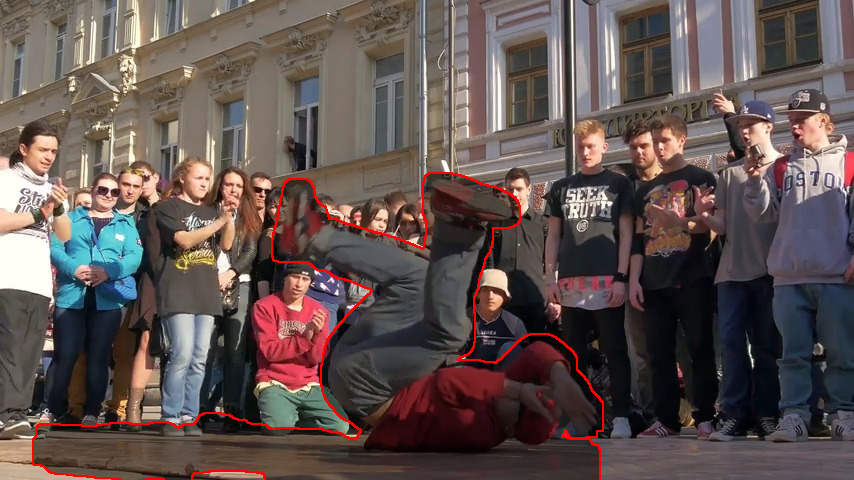} &
\includegraphics[align=c,width=\imwidth]{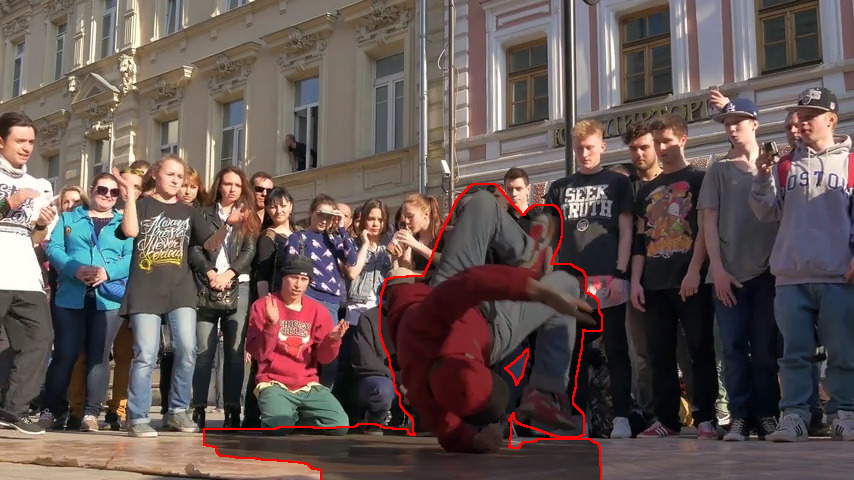} &
\includegraphics[align=c,width=\imwidth]{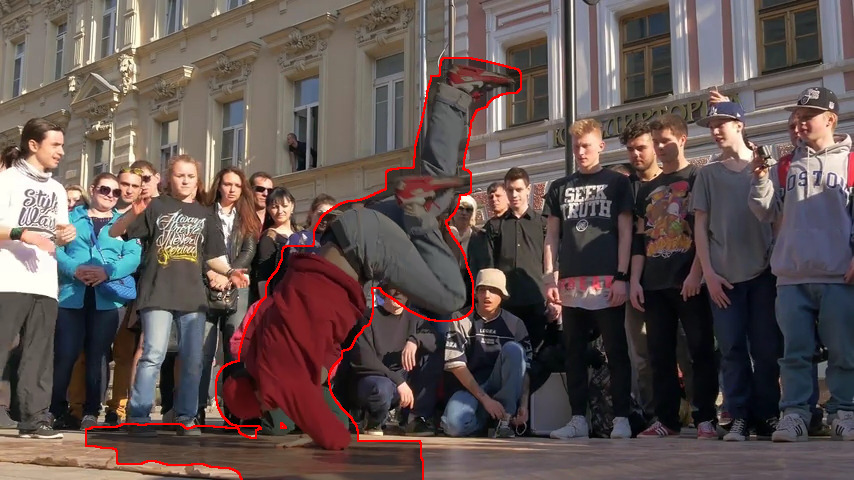} \\
Ours \scriptsize{14+prealign} &
\includegraphics[align=c,width=\imwidth]{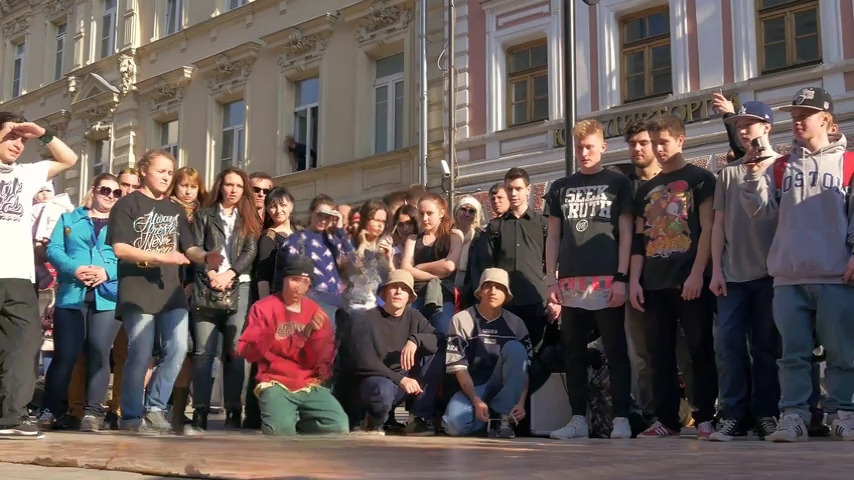} &
\includegraphics[align=c,width=\imwidth]{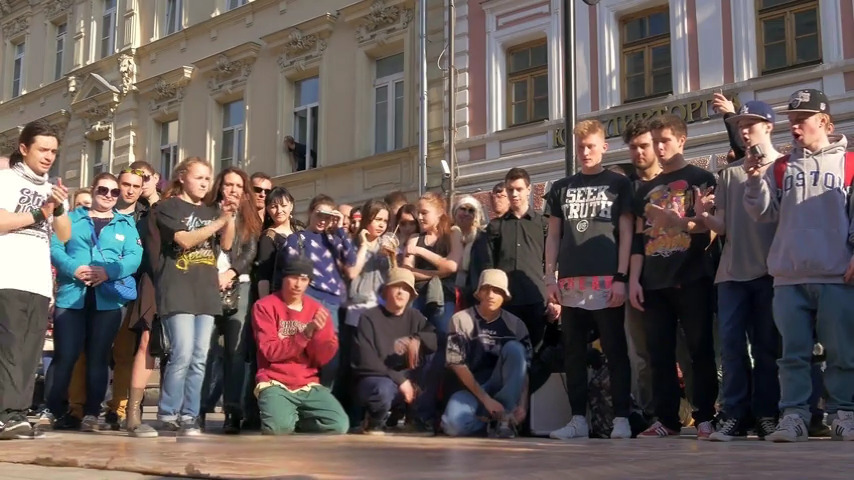} &
\includegraphics[align=c,width=\imwidth]{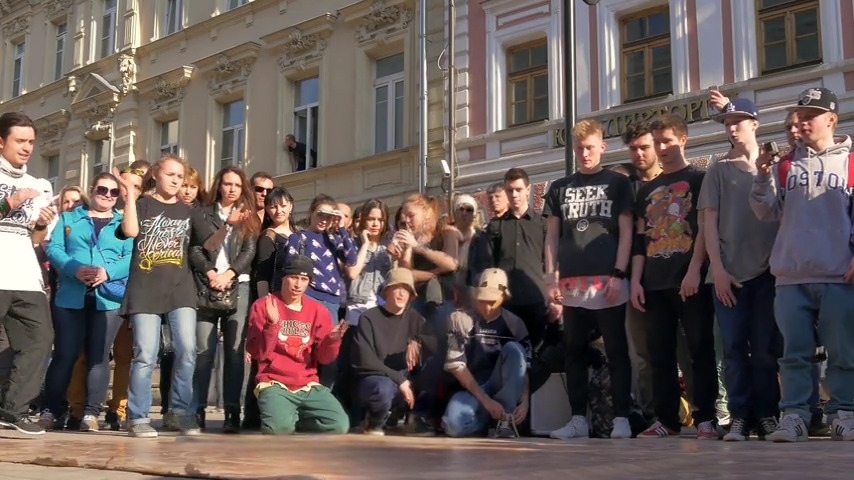} &
\includegraphics[align=c,width=\imwidth]{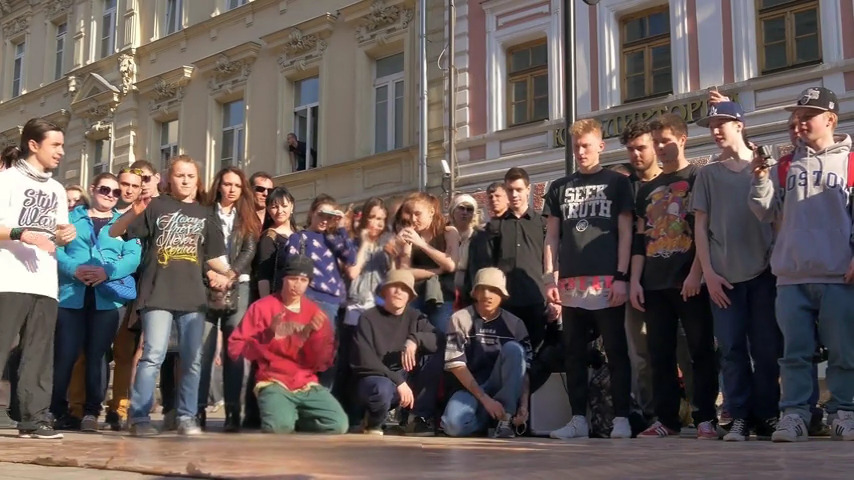} \\
Ours \scriptsize{14} &
\includegraphics[align=c,width=\imwidth]{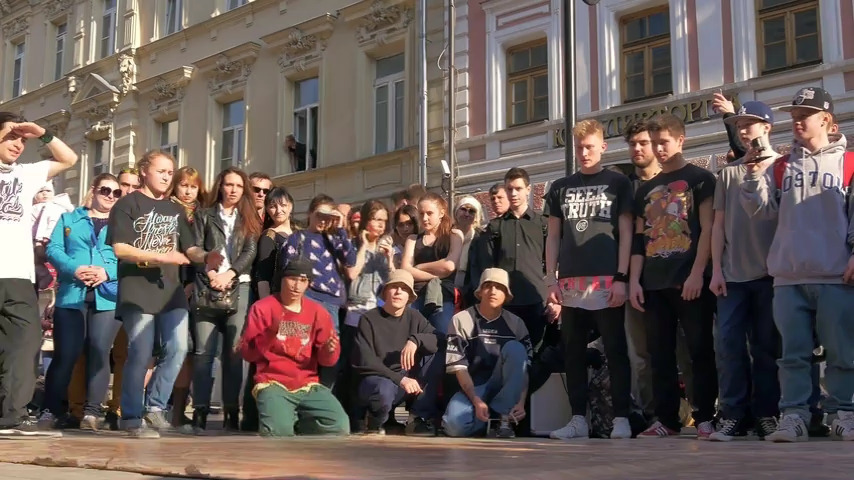} &
\includegraphics[align=c,width=\imwidth]{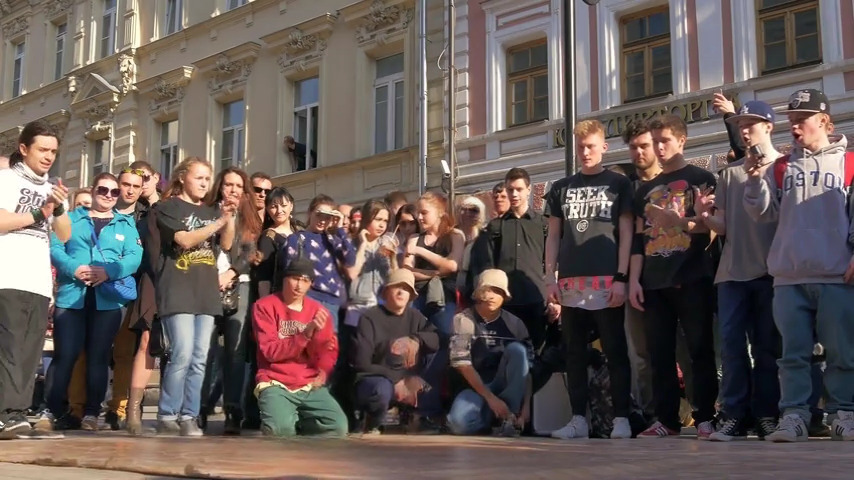} &
\includegraphics[align=c,width=\imwidth]{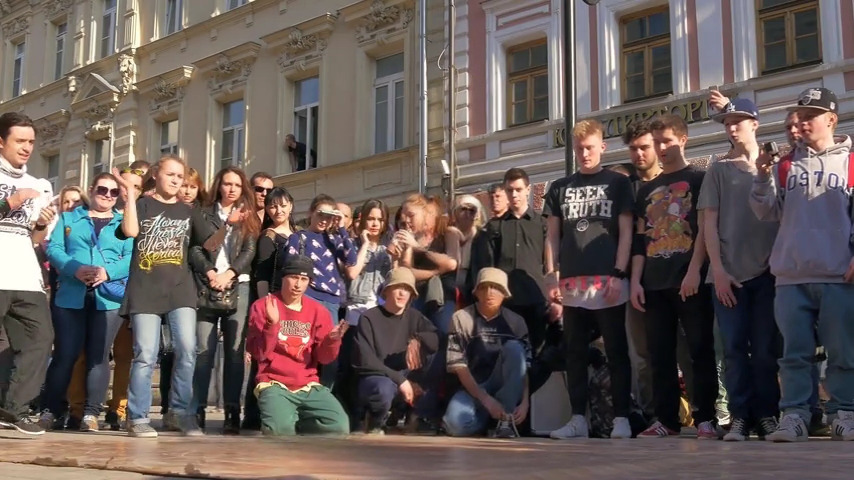} &
\includegraphics[align=c,width=\imwidth]{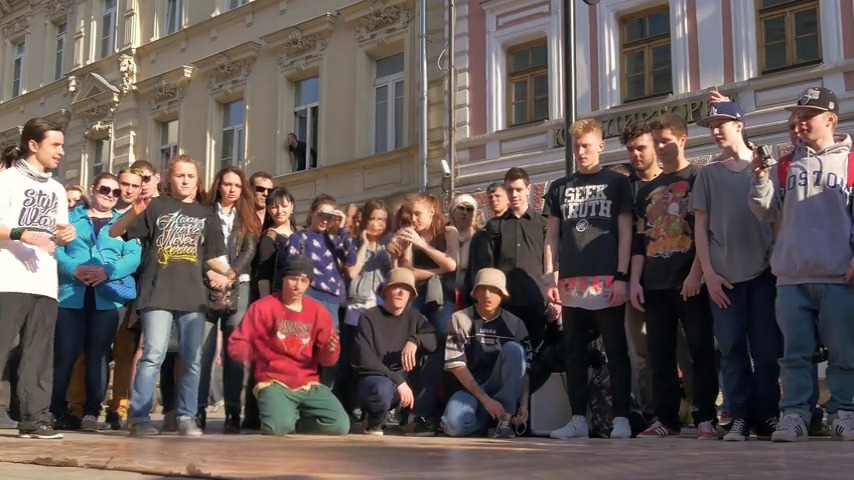} \\
LaMa\cite{lama} \scriptsize{+flow} &
\includegraphics[align=c,width=\imwidth]{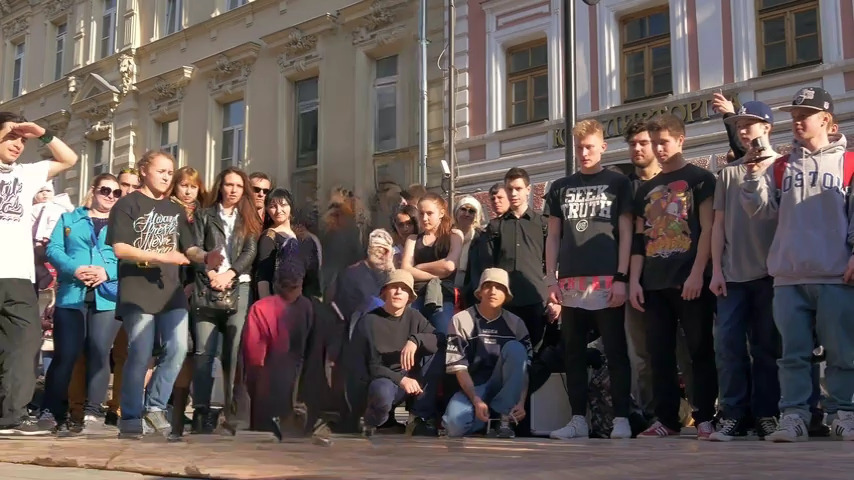} &
\includegraphics[align=c,width=\imwidth]{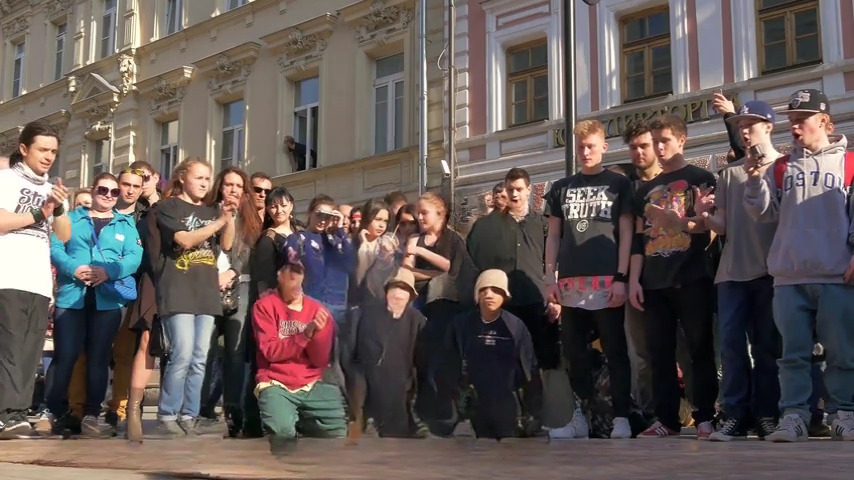} &
\includegraphics[align=c,width=\imwidth]{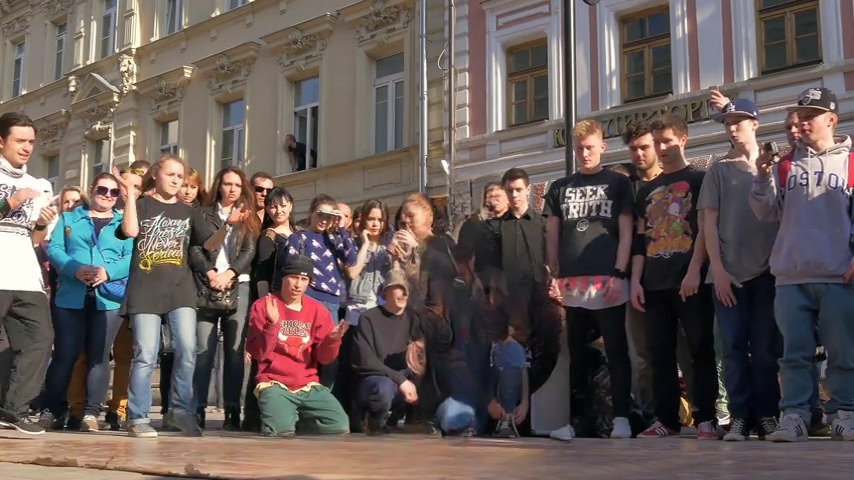} &
\includegraphics[align=c,width=\imwidth]{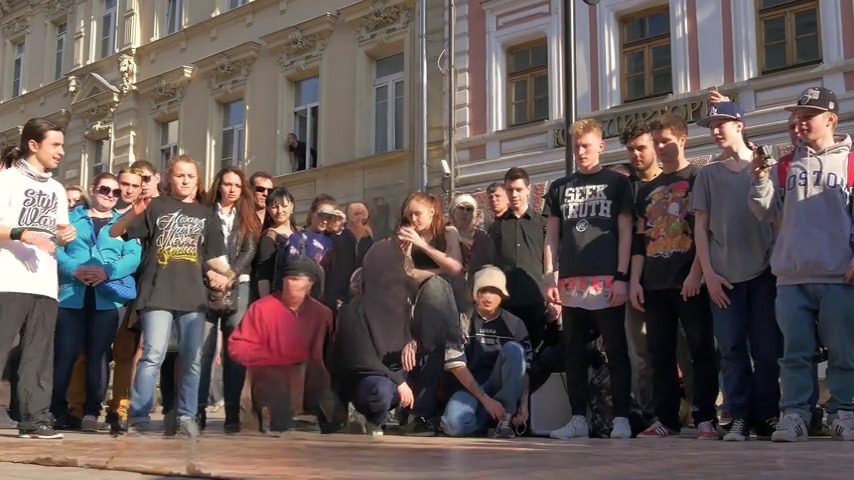} \\
DFCNet\cite{dfcnet} &
\includegraphics[align=c,width=\imwidth]{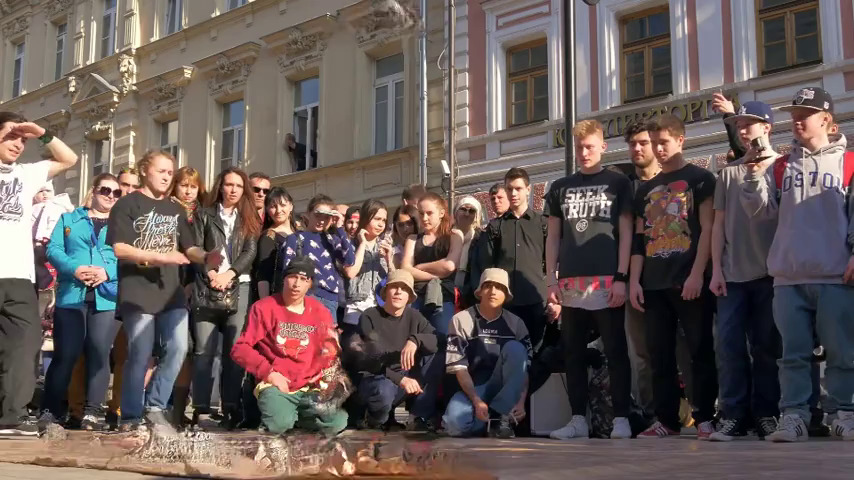} &
\includegraphics[align=c,width=\imwidth]{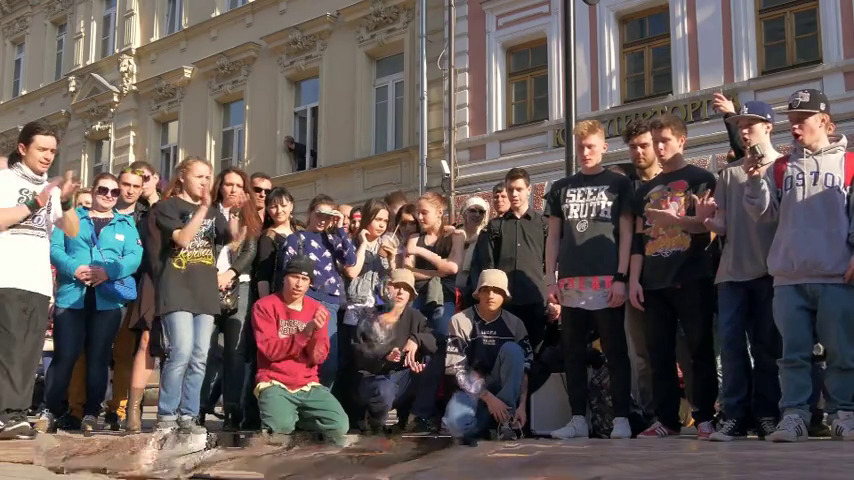} &
\includegraphics[align=c,width=\imwidth]{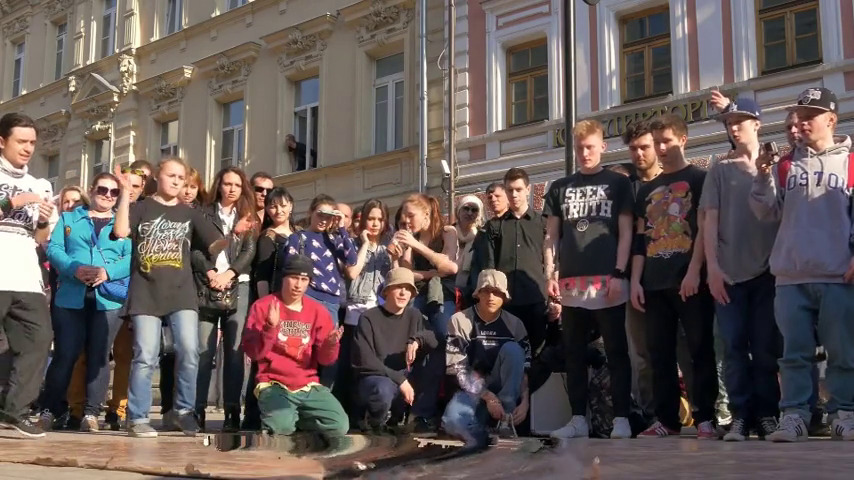} &
\includegraphics[align=c,width=\imwidth]{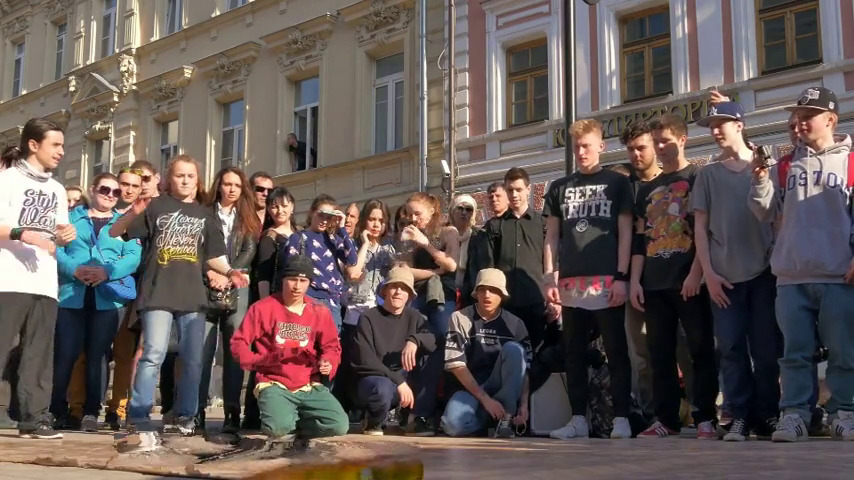} \\
E2FGVI\cite{e2fgvi} &
\includegraphics[align=c,width=\imwidth]{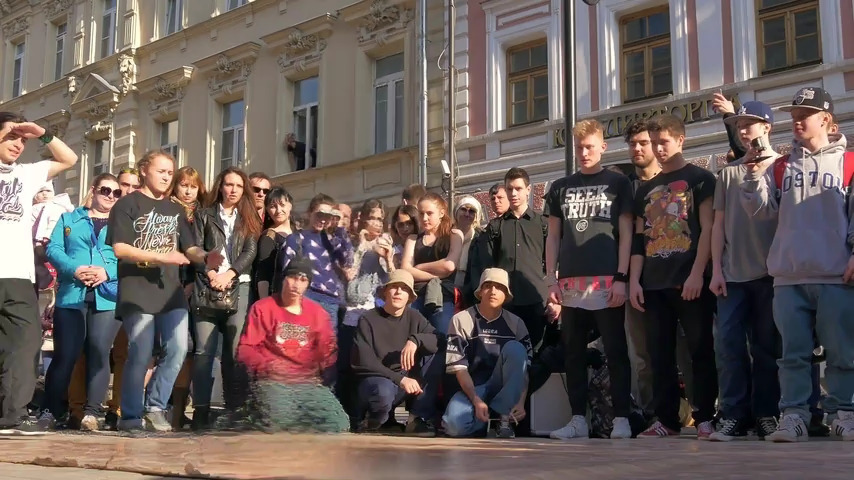} &
\includegraphics[align=c,width=\imwidth]{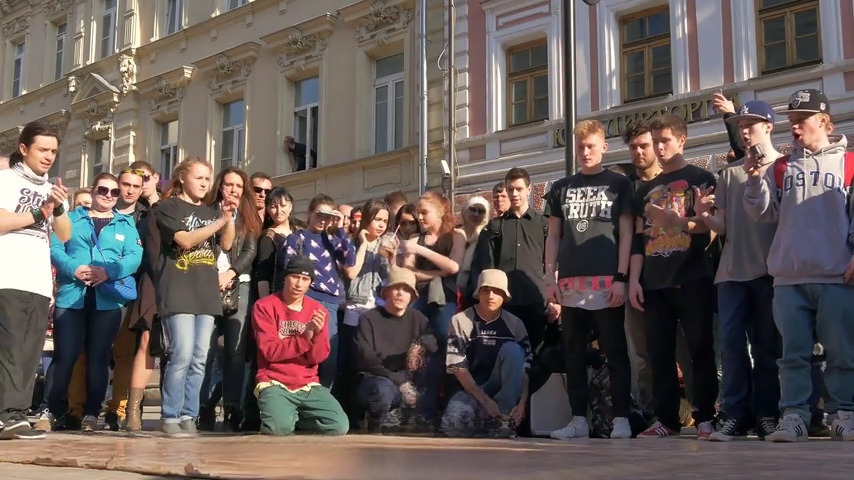} &
\includegraphics[align=c,width=\imwidth]{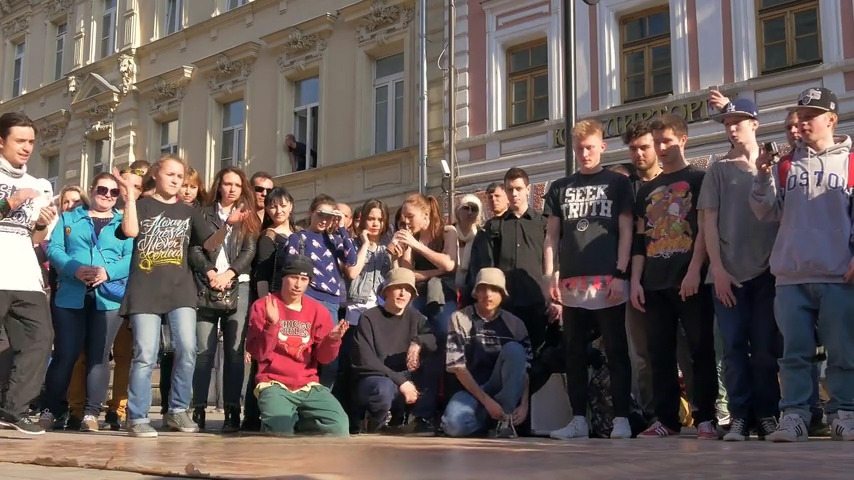} &
\includegraphics[align=c,width=\imwidth]{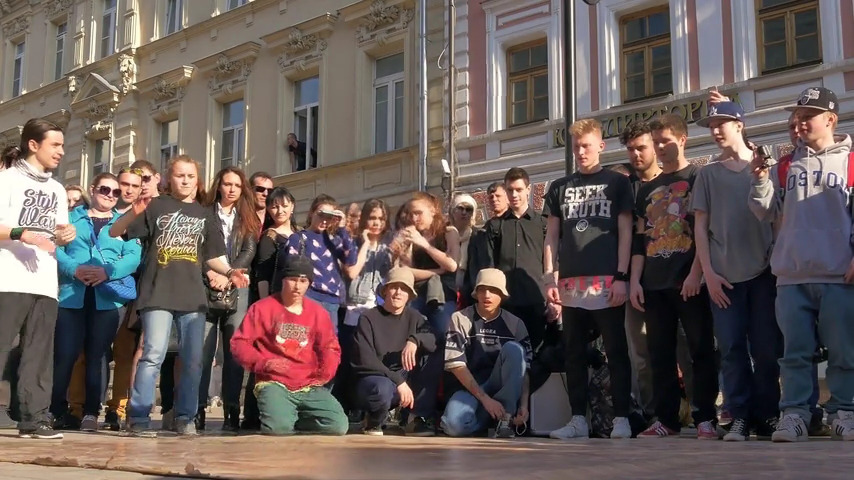} \\
FGVC\cite{fgvc} &
\includegraphics[align=c,width=\imwidth]{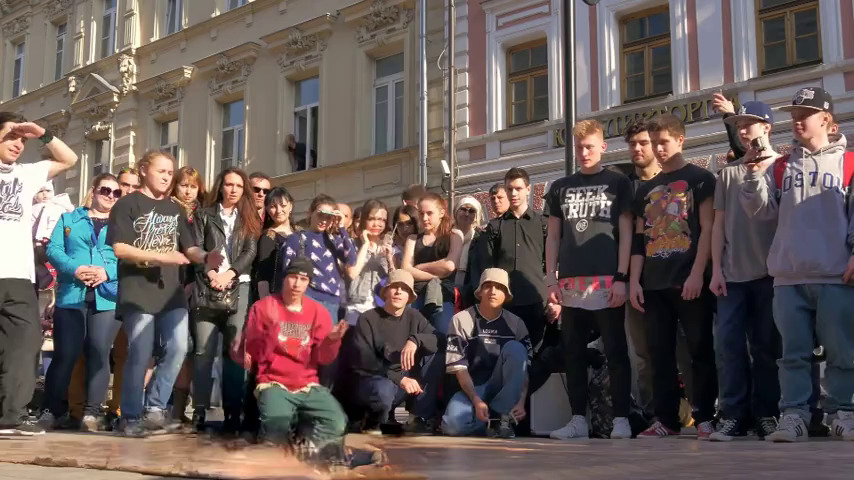} &
\includegraphics[align=c,width=\imwidth]{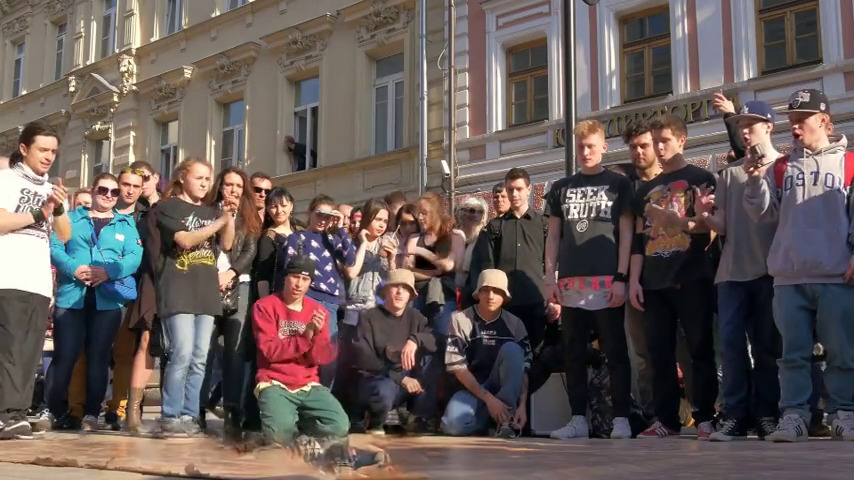} &
\includegraphics[align=c,width=\imwidth]{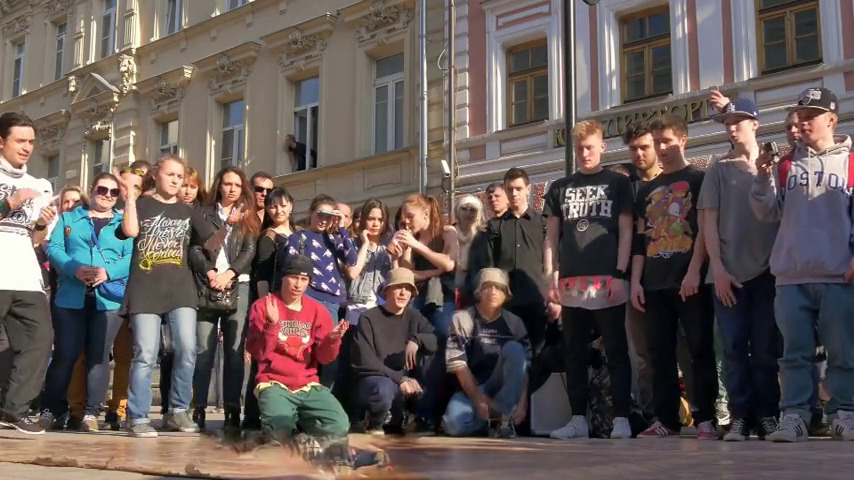} &
\includegraphics[align=c,width=\imwidth]{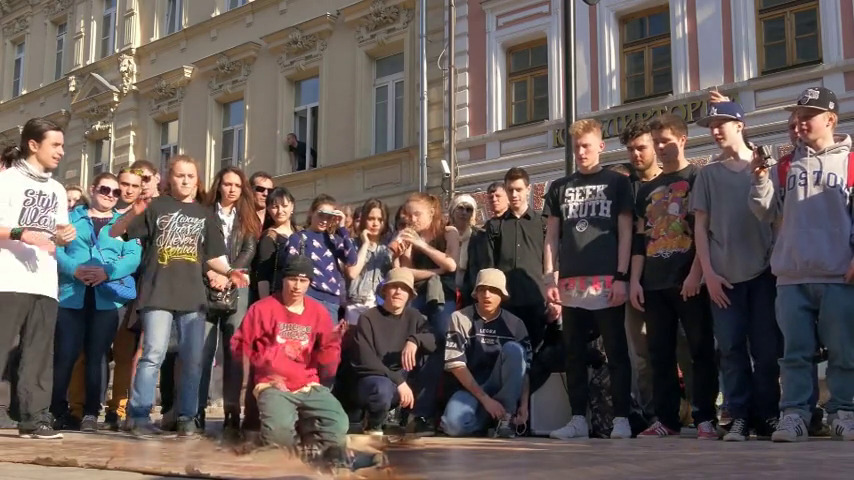} \\
	\bottomrule
  \end{tabular}
\caption{\label{fig:qualitativevibd} Qualitative results for video inpainting on
    the DAVIS\cite{davis} sequence \emph{breakdance}. Animated version at \url{\projecturl}.}
\end{figure*}
}

\newcommand{\tabinpainting}{
\begin{table}
\begin{footnotesize}
\begin{center}
 \begin{adjustbox}{max width=\linewidth}
  \begin{tabular}[t]{lrrr}
    \toprule
    \textbf{Method} & FID $\downarrow$ & LPIPS $\downarrow$ & SSIM $\uparrow$\\
    \midrule
    Transformer&15.19&0.173\footnotesize{$\pm$ 0.072}&0.814\footnotesize{$\pm$ 0.097}\\
    Ours w/o FFC&14.19&0.155\footnotesize{$\pm$ 0.075}&0.831\footnotesize{$\pm$ 0.096}\\
    LaMa&\underline{13.73}&\underline{0.147}\footnotesize{$\pm$ 0.077}&\underline{0.836}\footnotesize{$\pm$ 0.096}\\
    Ours&\textbf{13.15}&\textbf{0.145}\footnotesize{$\pm$ 0.074}&\textbf{0.837}\footnotesize{$\pm$ 0.095}\\
    \bottomrule
\end{tabular}
 \end{adjustbox}\vspace{-0.85em}
 \caption{\label{tabinpainting}Single Image Inpainting Results on Places\cite{places}.}
\end{center}\vspace{-2em}
\end{footnotesize}
\end{table}
}

\newcommand{\tabginpainting}{
\begin{table}
\begin{footnotesize}
\begin{center}
 \begin{adjustbox}{max width=\linewidth}
  \begin{tabular}[t]{lrrr}
    \toprule
    \textbf{Method} & FID $\downarrow$ & LPIPS $\downarrow$ & SSIM $\uparrow$\\
    \midrule
    LaMa + SpectralFuse&26.40&0.211\footnotesize{$\pm$ 0.115}&0.794\footnotesize{$\pm$ 0.138}\\
    Transformer&18.18&0.175\footnotesize{$\pm$ 0.094}&0.821\footnotesize{$\pm$ 0.128}\\
    Our w/o FFC&\underline{14.78}&\underline{0.140}\footnotesize{$\pm$ 0.082}&\textbf{0.851}\footnotesize{$\pm$ 0.110}\\
    Ours&\textbf{14.29}&\textbf{0.137}\footnotesize{$\pm$ 0.080}&\underline{0.848}\footnotesize{$\pm$ 0.110}\\
    \midrule
    Ours (big)&13.09&0.130\footnotesize{$\pm$ 0.074}&0.865\footnotesize{$\pm$ 0.101}\\
    \bottomrule
\end{tabular}
 \end{adjustbox}\vspace{-0.85em}
 \caption{\label{tabginpainting}Guided Image Inpainting Results on Places\cite{places}.}
\end{center}\vspace{-2em}
\end{footnotesize}
\end{table}
}

\newcommand{\tabvinpainting}{
\begin{table}
\begin{center}
 \begin{adjustbox}{max width=\linewidth}
  \begin{tabular}[b]{lrrrrrr}
    \toprule
    \textbf{Method} & FID $\downarrow$ & LPIPS $\downarrow$ & SSIM $\uparrow$ & VFID $\downarrow$ & PVCS $\downarrow$ & PCons $\uparrow$\\
    \midrule
    Ours \scriptsize{14+prealign}& \textbf{4.34} & \textbf{0.0034} & \textbf{0.9630} & 0.0543 & \underline{0.2263} & 42.39 \\
    Ours \scriptsize{14}& \underline{4.45} & \underline{0.0042} & \underline{0.9597} & \underline{0.0542} & 0.2339 & 41.45 \\
    Ours \scriptsize{6}& 4.68 & 0.0051 & 0.9581 & 0.0561 & 0.2443 & 41.14 \\
    LaMa\cite{lama} \scriptsize{+flow} & 5.28 & 0.0082 & 0.9459 & 0.0877 & 0.3151 & 40.22 \\
    DFCNet\cite{dfcnet} & 7.71 & 0.0054 \scriptsize{(0.0056)} & 0.9500 & 0.0627 & \textbf{0.2240} & \textbf{57.80} \\
    E2FGVI\cite{e2fgvi} & 8.22 & 0.0046 & 0.9585 & 0.0613 & 0.2519 & 33.19 \\
    FGVC\cite{fgvc} & 9.76 & 0.0065 \scriptsize{(0.0054)} & 0.9548 & 0.0564 & 0.2322 & 38.26 \\
    JointOpt\cite{jointopt} & 7.30 & 0.0058 \scriptsize{(0.0059)} & 0.9559 & \textbf{0.0530} & 0.2324 & 37.19 \\
    OPN\cite{opn} & 8.71 & 0.0062 \scriptsize{(0.0057)} & 0.9525 & 0.0708 & 0.2874 & 35.09 \\
    CPNet\cite{cpnet} & 13.21 & 0.0056 \scriptsize{(0.0068)} & 0.9574 & 0.0969 & 0.3048 & 43.91 \\
    STTN\cite{sttn} & 10.89 & 0.0083 \scriptsize{(0.0065)} & 0.9542 & 0.0911 & 0.3149 & 41.12 \\
    VINet\cite{vinet} & 23.63 & 0.0090 \scriptsize{(0.0084)} & 0.9481 & 0.1488 & 0.4312 &
    \underline{50.07} \\
    \bottomrule
\end{tabular}
 \end{adjustbox}\vspace{-0.85em}
 \caption{\label{tabvinpainting}Video Inpainting Result on DEVIL\cite{devil}.}
\end{center}\vspace{-2.5em}
\end{table}
}

\newcommand{\tabginpaintingtwo}{
  \begin{table}[h]
\begin{footnotesize}
\begin{center}
 \begin{adjustbox}{max width=\linewidth}
  \begin{tabular}[t]{lrrr}
    \toprule
    \textbf{Method} & FID $\downarrow$ & LPIPS $\downarrow$ & SSIM $\uparrow$\\
    \midrule
    LaMa + SpectralFuse&26.74&0.212\footnotesize{$\pm$ 0.116}&0.759\footnotesize{$\pm$ 0.139}\\
    Transformer&18.87&0.181\footnotesize{$\pm$ 0.096}&0.802\footnotesize{$\pm$ 0.130}\\
    Our w/o FFC&\underline{17.62}&\underline{0.157}\footnotesize{$\pm$ 0.091}&\textbf{0.833}\footnotesize{$\pm$ 0.118}\\
    Ours&\textbf{16.90}&\textbf{0.154}\footnotesize{$\pm$ 0.089}&\underline{0.834}\footnotesize{$\pm$ 0.118}\\
    \midrule
    Ours (big)&15.45&0.148\footnotesize{$\pm$ 0.084}&0.849\footnotesize{$\pm$ 0.111}\\
    \bottomrule
\end{tabular}
 \end{adjustbox}\vspace{-0.85em}
 \caption{\label{tabginpaintingtwo}Guided Image Inpainting Results on Places\cite{places} using two keyframes.}
\end{center}\vspace{-2em}
\end{footnotesize}
\end{table}
}

\newcommand{\projecturl}{https://github.com/runwayml/guided-inpainting}

\newcommand{\RR}{\mathbb{R}}

\begin{document}

\title{Towards Unified Keyframe Propagation Models}

\author{%
  Patrick Esser$^1$ \qquad Peter Michael$^{1,2}$ \qquad Soumyadip Sengupta$^2$ \\
  \small{$^1$\href{https://research.runwayml.com/}{Runway ML} \qquad $^2$\href{https://www.washington.edu/}{University of Washington}}\\
  \url{\projecturl}
}

\twocolumn[{%
\maketitle%
\figone%
}]

\begin{abstract}
  Many video editing tasks such as rotoscoping or object removal require the
  propagation of context across frames. While transformers and other
  attention-based approaches that
  aggregate features globally have demonstrated great success at
  propagating object masks from keyframes to the whole video, they
  struggle to propagate high-frequency details such as textures faithfully.
  We hypothesize that this is due to an inherent bias
  of global attention towards low-frequency features.
  To overcome this limitation, we present a two-stream approach, where
  high-frequency features interact locally and low-frequency features interact
  globally.
  The global interaction stream
  remains robust
  in difficult situations such as large camera motions, where
  explicit alignment fails.
  The local interaction stream propagates high-frequency details
  through deformable feature aggregation and, informed by the global
  interaction stream, learns to detect and correct
  errors of the deformation field.
  We evaluate our two-stream approach for inpainting tasks, where experiments
  show that it improves both the propagation of features within a single
  frame as required for image inpainting, as well as their propagation from
  keyframes to target frames. Applied to video inpainting, our approach
  leads to $44\%$ and $26\%$ improvements in FID and LPIPS scores.
\end{abstract}

\section{Introduction}
\label{sec:introduction}
\figmodel

To satisfy the ever-increasing demand for video content, new solutions
for fast and simple video creation are required. By reducing the need for
tedious frame-by-frame operations, machine learning can help creators to focus
on the creative aspects of story-telling.
A promising approach to reduce manual work without limiting creativity is based
on keyframe propagation, where the desired result is specified on a few
frames and automatically propagated to the entire video.
This formulation and the need to propagate context across
frames is common to many video editing tasks. For example,
to separate foreground objects from the background, object selection masks must be
propagated,
to stylize videos, artistic edits of frames must be propagated,
and to remove objects from videos, the background must be propagated from
frames where it is visible (see Fig.~\ref{figone}).

Given this commonality, it is natural to ask whether a single unified approach
can handle all of these tasks.
A promising candidate for such a solution is a transformer \cite{transformer} based architecture. Not
only has it been a driving force of consolidation across a
wide range of tasks and modalities \cite{karpathy,perceiver}, but, by relying
on global, affinity-weighted feature aggregation, it has the potential to
propagate context between multiple non-aligned frames. Furthermore,
transformers and related approaches that globally aggregate features have been successfully applied to object
mask propagation \cite{Oh2019VideoOS, stcn}.

However, Fig.~\ref{figone} shows that a direct adaptation of transformers
to keyframe-guided inpainting for object removal fails to
propagate details about the background from keyframes. This
observation is in agreement with the findings of \cite{park2022}, that
global feature aggregation via attention acts as a low-pass filter, which
hinders propagation of high-frequency details. Based on this,
we present a two-stream model consisting of a stream of locally interacting
features (LIF) and a stream of globally interacting features (GIF). To avoid
loss of high-frequency details, LIF operates on a high-resolution
representation and uses local convolutional operations, which exhibit the
opposite behavior of attention and act as high-pass-filters \cite{park2022}.
High-frequency details are then propagated with a deformable
feature aggregation between LIF streams of different frames, whereas
robust low-frequency features are propagated with attention in the GIF
stream.
(see Fig.~\ref{fig:model}). Experiments demonstrate that this design improves
propagation both within a single frame, evaluated via image inpainting,
as well as across frames, evaluated via guided image inpainting and video inpainting.

\section{Background}
\label{sec:relatedwork}

\textbf{Transformers in Vision} Following the success of transformers in
natural language processing \cite{transformer,gpt}, their architecture has been widely adopted
for vision tasks \cite{vits,vitsurvey}. While the attention mechanism is permutation invariant
and can model all pairwise feature interactions, its quadratic complexity with
respect to input length requires adaptions for high-dimensional image data \cite{perceiver}.
Many vision-specific modifications such as windowed attention \cite{swin} introduce
assumptions about possible interactions and thus require aligned input data,
making them unsuitable for keyframe propagation.
Other adaptations are based on approximative attentions mechanisms
\cite{linformer,performer} or
attention on compressed representations \cite{DBLP:journals/corr/abs-2103-03841,vqgan} which
introduces a bottleneck for propagating features. Even without such
bottlenecks, recent work \cite{park2022} shows that the attention mechanism
is inherently biased towards low-frequency features. Hence, we aim to better
understand what is required for faithful propagation of both low- and
high-frequency content in order to get closer to a unified approach for
keyframe propagation.%

\textbf{Frame Propagation} Propagation of information such as color or object
masks across video frames can be formulated as optimization problems
\cite{Agarwala2004KeyframebasedTF,Levin2004ColorizationUO}.  Since these lead
to slow runtimes, deep learning-based approaches have been explored as faster
alternatives \cite{Zhang2017RealtimeUI,Meyer2018DeepVC} or as more accurate but
slower alternatives \cite{Caelles2017OneShotVO} by fine-tuning models per
video. Instead of relying on a sequential forward propagation through the
video, space-time memory networks \cite{Oh2019VideoOS}, which bear resemblance
to single-layer transformers, enable propagation of object masks from arbitrary
keyframes.

\textbf{Inpainting} Patch-based synthesis
\cite{Wexler2007SpaceTimeCO,Barnes2009PatchMatchAR} provides an elegant framework that
demonstrates how both image- and video-inpainting can be formulated as
propagation problems. For improved runtimes and to benefit from learned data
priors, deep learning-based solutions have been proposed for image
\cite{lama,comodgan,aotgan,regionwise,deepfillv2,edgeconnect} and
video inpainting \cite{fgvc,dfcnet,jointopt,opn,cpnet,sttn,vinet}.
Most video inpainting approaches rely on sequential forward prediction in time,
which limits their applicability in interactive scenarios. Closest to our work
on keyframe-guided inpainting are \cite{transfill} and \cite{geofill}, but both
rely on accurate estimation of scene geometry for alignment between frames,
which limits applications to static scenes \cite{geofill}. In contrast, we only
utilize a very rough alignment based on optical flow which is interpolated in
masked areas. While this introduces errors and deformations, our model can
detect and correct them based on local and global feature interactions (see
Fig.~\ref{figqualinpaint}).

\figqualinpaint

\section{Approach}
\label{sec:approach}
\newcommand{\tin}{\text{in}}
\newcommand{\tgt}{\text{tgt}}
\newcommand{\src}{\text{src}}
\newcommand{\xtgt}{x_{\text{tgt}}}
\newcommand{\xsrc}{x_{\text{src}}}
\newcommand{\model}{f}
\newcommand{\fenc}{f_{\text{enc}}}
\newcommand{\fdec}{f_{\text{dec}}}
\newcommand{\hgif}[1]{h_{\text{GIF}}^{#1}}
\newcommand{\hlif}[1]{h_{\text{LIF}}^{#1}}
\newcommand{\readop}{\text{read}}
\newcommand{\writeop}{\text{write}}
\newcommand{\softmax}{\text{softmax}}
\newcommand{\sigmoid}{\text{sigmoid}}

Our goal is to develop a model that can propagate information from keyframes to
target frames. Let $\xtgt \in \RR^{H_\tin \times W_\tin \times C_\tgt}$ be the target frame
and $\xsrc \in \RR^{T \times H_\tin \times W_\tin \times C_\src}$ be a sequence of
keyframes. For inpainting, $C_\tgt = C_\src = 4$, containing RGB intensities of
masked frames and the mask. The completion of the
target frame depends on unmasked content from both $\xtgt$ and $\xsrc$.

\subsection{Feature Encoding and Decoding}

\textbf{LIF Encoding}
The main goal of the LIF stream is to preserve and
propagate high-frequency details. To initialize this stream, we follow
previous works and use a strided convolutional architecture to extract features
from the inputs. By keeping the stride $s$ relatively small while expanding the
number of feature channels $c$, (in practice we use a stride $s=8$ with $c=512$
channels), these features, together with a similar deconvolutional
architecture, provide\phantom{s}
an almost lossless feature encoding.
We encode each keyframe and the target frame individually and model interactions only
afterwards. Thus, every frame
represented in the LIF stream by $\hlif{0} \in \RR^{H \times W \times c}$,
where $H=\frac{H_\tin}{s}$ and $W=\frac{W_\tin}{s}$.

\textbf{GIF Initialization}
The goal of the GIF stream is to model robust low-frequency interactions within
and across frames.
We initialize this stream with a set of learned
parameters $\hgif{0} \in \RR^{M \times d}$, where $M$ is the sequence length
and controls the
computational complexity of the stream. The learned parameters serve as a positional
encoding.

\textbf{Decoding}
The encoding stage results in two feature representations,
$\hlif{0}, \hgif{0} = \fenc(x)$ for each frame $x$. These will be
processed by a sequence of $L$ blocks which produce $\hlif{L}, \hgif{L}$.
The final output is then predicted from the $\hlif{L}$ representation of the
target frame, i.e. $y=\fdec(\hlif{L})$. Next, we describe how the blocks map
$\hlif{i}, \hgif{i} \mapsto \hlif{i+1}, \hgif{i+1}$.

\subsection{Intra-Frame and Inter-Stream Interactions}\vspace{-0.5em}
The LIF stream consists of convolutional residual blocks for local
interaction, while the GIF stream consists of attention blocks for global
interactions.
Read and write operations exchange complementary features between them.

\textbf{Read Operation}
Since generally $M < N \coloneqq H \cdot W$, we have to average LIF features or
select a subset of them to read into the GIF stream.
We use a read module that learns how to read
features from the LIF into the GIF stream.

We assume that $M=m^2$ is a square number and that $H \equiv 0 \mod m$ and $W
\equiv 0 \mod m$ (otherwise interpolating to the closest
height and width that is divisible by $m$). We divide the LIF feature map
into $M$ equal-sized patches and learn which features of the $j$-th patch
to read into the $j$-th GIF feature by
forming an importance weighted sum of
projected features. If $P_j \in
\RR^{(\frac{H}{m}\cdot\frac{W}{m})\times c}$ is the $j$-th patch,
\begin{equation}
  \readop(P_j) = \softmax(W_S P_j^t) P_j W_V,\vspace{-0.5em}
\end{equation}
where $W_S \in \RR^{1 \times c}$ and $W_V \in \RR^{c \times d}$ are learnable
weights.
We follow common design choices of
attention and (i) use multiple read heads with their own score and
value projections and (ii) apply a Layer Normalization followed by a linear
projection and a residual connection on the GIF stream.

\textbf{Write Operation}
After processing the GIF stream with attention blocks, we write the result back
into the high-capacity LIF stream. We use a similar approach as for the read
operation, except that the attention weights are now computed on the transposed
patch and values are computed from the GIF stream. If
$P_j \in \RR^{(\frac{H}{m}\cdot\frac{W}{m})\times c}$ is the $j$-th LIF patch
and $G_j \in \RR^{1 \times d}$ is the $j$-th GIF feature,
\begin{equation}
  \writeop(P_j, G_j) = \sigmoid(P_j W_S^t) G_j W^t_V,\vspace{-0.5em}
\end{equation}
with $W_S \in \RR^{1 \times c}$ and $W_V \in \RR^{c \times d}$. As in
the read operation, we use multiple write heads
followed
by layer norm, projection, and a residual connection on the LIF stream.

Together, a read operation, local and global interactions in the
LIF and GIF streams, followed by a write operation represent one intra-frame
block as shown in Fig.~\ref{fig:model}.

\vspace{-0.5em}
\subsection{Cross-Frame Interactions}\vspace{-0.5em}
\enlargethispage{\baselineskip}

\textbf{Global Cross-Frame Propagation}
To propagate information from keyframes to the target, we replace
self-attention in the GIF stream with cross-attention.
We use a symmetric design, where the same
block is used for self-attention on keyframes and for cross-attention
on target frames, by concatenating keys and values from keyframes.
If $g_\src \in \RR^{M \times d}$ and $g_\tgt \in \RR^{M \times
d}$ denote GIF features from a keyframe and the target frame, respectively,
we compute the attention operation $A$ for keyframes and target as
\begin{align*}
  \text{A}(g_\src, \varnothing) &= \softmax(g_\src W_Q^t (g_\src W_K^t)^t )
  (g_\src W_V) \\
  \text{A}(g_\tgt, g_\src) &= \softmax(g_\tgt W_Q^t
  ([g_\tgt, g_\src] W_K^t)^t )
  [g_\tgt, g_\src] W_V,
\end{align*}
with $W_Q \in \RR^{d_k \times d}$, $W_K \in \RR^{d_k \times d}$, and $W_V \in \RR^{d \times d}$.

\textbf{Deformable Write Operation}
To overcome limitations of attention-based propagation of high-frequency features,
we introduce a complementary approach based on an explicit alignment with a
deformable write module that fuses LIF features from keyframes into those of
the target frame.

Let $\hlif{i} \in \RR^{H \times W \times c}$ denote LIF features of keyframe
$i$, and
$\phi_i \in \RR^{H \times W \times 2}$ a flow field such that a position $p$ in
the target frame corresponds to the position $p + \phi_i(p)$ in the $i$-th
keyframe. We write $[\hlif{i}(p + \phi_i(p))] \in \RR^{T \times c}$ to denote the
stacked, bilinearly interpolated values of $\hlif{i}$
at positions $p+\phi_i(p)$.
The deformable write operation then aggregates these interpolated features
based on confidence scores,\vspace{-0.5em}
\begin{equation*}
  \sum \softmax([\hlif{i}(p+\phi_i(p))] W_Q^T) \odot [\hlif{i}(p+\phi_i(p))] W_V\vspace{-0.5em}
\end{equation*}
where $\odot$ is elementwise multiplication, $W_Q \in \RR^{c \times c}$ and
$W_V \in \RR^{c \times c}$ are learnable parameters, and the sum and softmax
are applied along the stacked dimension. The result is added to the
LIF feature stream of the target frame.

In contrast to global cross-frame interaction, this operation
can propagate high-frequency details but it is also more
susceptible to errors due to wrongly estimated flow fields $\phi_i$. For
the estimation of confidence scores, we
thus concatenate forward-backward consistency values of the flow
field. The flow fields are estimated with a
pre-trained optical flow model \cite{raft} operating at a fixed resolution of
$256 \times 256$ and interpolated to the spatial size of the LIF features.
\enlargethispage{\baselineskip}

\tabinpainting
\tabginpainting
\tabvinpainting

\vspace{-0.5em}
\section{Experiments}\vspace{-0.5em}
We evaluate the propagation capabilities of our model
on different inpainting tasks. Image inpainting requires context propagation
within a frame, while guided inpainting and video inpainting require context propagation across frames to predict plausible and consistent content in masked areas.

We use LaMa \cite{lama}, which relies on Fast Fourier Convolutions (FFCs) to
take global context into account, as a strong baseline and
follow its training and evaluation protocol based on the Places dataset
\cite{places}. Tab.~\ref{tabinpainting} compares the inpainting performance of
different approaches under the same computational budget, which is calibrated
through the number of blocks. A transformer
approach, which is similar to our model without the LIF stream, performs worst,
demonstrating its difficulties to faithfully propagate context. Our approach
with spatial convolutions in the LIF stream (Ours w/o FFC) improves
performance but achieves worse performance than LaMa. However, in combination
with FFCs in the LIF stream (Ours), our model achieves the best performance,
suggesting complementary properties of spectral convolutions and propagation
capabilities.

To evaluate context propagation across frames, we generate two synthetic
keyframes for each example by applying random
transformations, deformations, and masks.
We further investigate propagation in the spectral
domain and include a variant of LaMa that aggregates Fourier features across
frames (Lama+SpectralFuse). Both the quantitative results in
Tab.~\ref{tabginpainting}, as well as the qualitative results in
Fig.~\ref{figqualinpaint} show that this fails to propagate context unless a
keyframe happens to be aligned with the target frame (second row in
Fig.~\ref{figqualinpaint}). The transformer-based approach can take
keyframe-context into account but fails to propagate high-frequency details.
Our approach performs significantly better than these two baselines both with
and without FFCs, which demonstrates the improved propagation capabilities.

Finally, we evaluate how well our approach generalizes to real
videos. We train a bigger variant of our best performing model for guided
inpainting (Ours (big)), and use it to generate every 20th frame of a masked
video.
The remaining frames are propagated based on
optical flow. We use the DEVIL evaluation \cite{devil} to compare
different approaches in Tab.~\ref{tabvinpainting}.
The subscript indicates the number of keyframes used.
We also include results obtained by propagating frames
inpainted without context as \emph{Lama\cite{lama}{\scriptsize{+flow}}}.
Overall, we observe increasing performance with an increasing number of
keyframes available, which demonstrates that our approach is able to
propagate context across frames and that this leads to improved video
inpainting performance.
\enlargethispage{\baselineskip}

\vspace{-0.5em}
\section{Conclusion}\vspace{-0.5em}
We propose a novel approach for keyframe propagation based on a two-stream
approach, where one stream models the local interactions of high-frequency
features and the other models the global interactions of low-frequency
features. We show that this method improves image and video
inpainting performance. We are investigating applications of this
architecture to other tasks such as object segmentation and matting, with the
goal to find a unified approach.

\clearpage
\FloatBarrier
{\small
\bibliographystyle{ieee_fullname}
\bibliography{ms}
}

\newpage
\FloatBarrier
\onecolumn
\appendix
\begin{center}
\LARGE\textbf{Supplementary Materials}
\end{center}
We include additional details on the evaluations in Sec.~\ref{sec:evaldetails}
and additional qualitative results in Sec.~\ref{sec:qualresults}.
Implementation details and video results can be found at \url{\projecturl}.

\section{Evaluation Details}
\label{sec:evaldetails}
\textbf{(Guided) Image Inpainting}
In Tab.~\ref{tabinpainting} and \ref{tabginpainting}, we use a
subset of 2k images from the Places\cite{places} validation split at a resolution
of $256\times 256$ pixels. \emph{Ours (big)} uses two cross-frame blocks and 12
intra-frame blocks, all other models use two cross-frame blocks and four intra-frame
blocks. We train models with a learning rate of $3.2 \cdot 10^{-4}$ and an
effective batch size of 32, accumulated over two batches and distributed across
two V100 GPUs.

For guided inpainting, we use two keyframes during training and four for the evaluation in
Tab.~\ref{tabginpainting}. Evaluation results with two keyframes can be
found in Tab.~\ref{tabginpaintingtwo}. The performance of all
methods improves as context from more keyframes becomes available and their
relative performance-ordering to one another remains the same.

\textbf{Video Inpainting}
For the evaluation in Tab.~\ref{tabvinpainting}, we use the evaluation
code\footnote{\url{https://github.com/MichiganCOG/devil}} provided by the DEVIL
benchmark \cite{devil}. We use the `flickr-all` video subset combined with the
`fvi-fgs-h` mask subset which contains large inpainting masks and represents a
challenging setting. For 
    DFCNet\cite{dfcnet},
    FGVC\cite{fgvc},
    JointOpt\cite{jointopt},
    OPN\cite{opn},
    CPNet\cite{cpnet},
    STTN\cite{sttn} and
    VINet\cite{vinet} we use the pre-computed inpainting results on
    `flickr-all\_fvi-fgs-h` provided by \cite{devil} and recompute the metrics.
    This reproduces the results reported in \cite{devil} except for the LPIPS
    metrics, for which we include the results from \cite{devil} in parentheses.

In addition, we compute inpainting results from concurrent video inpainting
work E2FGVI\cite{e2fgvi} using their provided
code\footnote{\url{https://github.com/MCG-NKU/E2FGVI}} and \emph{E2FGVI-HQ}
checkpoint.

The \emph{Ours} entries in Tab.~\ref{tabvinpainting} always refer to results
computed with the \emph{Ours (big)} checkpoint that contains 12 intra-frame
blocks. The number in the subscript indicates the number of keyframes. For $6$,
we use frames with offsets $\pm\{10, 20, 40\}$, and for $14$, we use frames
with offsets $\pm\{5, 10, 15, 20, 40, 60, 80\}$ to capture more context from
the video. The subscript {\scriptsize{+prealign}} refers to an extra step of
manually fusing context from other frames in areas where the optical flow is
reliable. This brings some additional gains and hints that there is still room
for improvement regarding the propagation of high-frequency details from other
frames. We hypothesize that deformable writes on higher-resolution feature maps
and/or the use of skip connections in the architecture could further improve
the propagation capabilities.

To evaluate the importance of taking context from keyframes into
account for video inpainting, we also include
\emph{LaMa\cite{lama}\scriptsize{+flow}}, which uses the \emph{big-lama}
checkpoint,
provided\footnote{\url{https://github.com/saic-mdal/lama}} by the authors of
\cite{lama}, in combination with the same flow-propagation of intermediate
frames as for the \emph{Ours} entries. This represents a strong non-guided
baseline for video inpainting which can achieve relatively good FID scores but
fails to take the context from the video into account as indicated by bad LPIPS
scores. With the ability to take context from other frames into account, our
approach can achieve the best scores for both measures.

\tabginpaintingtwo

\section{Qualitative Results}
\label{sec:qualresults}
We include additional qualitative results for guided image inpainting in
Fig.~\ref{fig:suppinpaintingsamples} and \ref{fig:suppinpaintingsamplesalt}, and
a larger version of Fig.~\ref{figqualinpaint} in
Fig.~\ref{fig:suppfigqualinpaint}. For video inpainting, we show qualitative
results for DAVIS\cite{davis} sequences in
Fig.~\ref{fig:qualitativevibear}-\ref{fig:qualitativevibd}. Animated versions
for all DAVIS sequences can be found online at \url{\projecturl}. The results
of other methods for these sequences are taken from the online
supplementary\footnote{\url{https://filebox.ece.vt.edu/~chengao/FGVC/pages/object_removal.html}}
of \cite{fgvc}
when available and recomputed otherwise.

Note that the long-term temporal stability of our method is reduced compared to
other approaches, because we compute each chunk of $20$ frames from the video
independently. This gives computational speed advantages by
enabling parallelization over chunks as opposed to a fully sequential processing
required by other approaches. On shorter timescales, the animated versions
demonstrate that our approach exhibits significantly less short-term temporal
instabilities like flickering.

\figqualinpaintsuppalt
\figqualinpaintsupp
\suppfigqualinpaint
\figqualitativevibear
\figqualitativevibdf
\figqualitativevibd

\end{document}